\pdfoutput=1
\documentclass[11pt]{article}

\usepackage{ACL2023}

\usepackage{times} 
\usepackage{latexsym}
\usepackage{multirow}
\usepackage{booktabs}
\usepackage[T1]{fontenc}
\usepackage{pbox}
\usepackage[utf8]{inputenc}

\usepackage{microtype}

\usepackage{inconsolata}

\usepackage{mathtools}

\usepackage{enumitem}

\usepackage{xcolor}

\usepackage{url}

\usepackage{float}

\usepackage{algorithm}
\usepackage{algorithmic}

\usepackage{caption}
\title{Assessing LLMs for Zero-shot Abstractive Summarization Through the Lens of Relevance Paraphrasing}

\author{
Hadi Askari\footnotemark[2] $\,$, Anshuman Chhabra\footnotemark[4] $\,$, Muhao Chen\footnotemark[2] $\,$, Prasant Mohapatra\footnotemark[4]\\
\footnotemark[2] $\,$ Department of Computer Science, University of California, Davis\\%~\\
\footnotemark[4] $\,$ Department of Computer Science and Engineering, University of South Florida\\%~\\
\texttt{\{haskari,muhchen\}@ucdavis.edu, \{anshumanc,pmohapatra\}@usf.edu}%~\\
}

\begin{document}

\maketitle
\begin{abstract}
Large Language Models (LLMs) have achieved state-of-the-art performance at zero-shot generation of abstractive summaries for given articles. However, little is known about the robustness of such a process of zero-shot summarization.
To bridge this gap, we propose \textit{relevance paraphrasing}, a simple strategy that can be used to measure the robustness of LLMs as summarizers. The relevance paraphrasing approach identifies the most \textit{relevant} sentences that contribute to generating an ideal summary, and then \textit{paraphrases} these inputs to obtain a minimally perturbed dataset. Then, by evaluating model performance for summarization on both the original and perturbed datasets, we can assess the LLM's one aspect of robustness. We conduct extensive experiments with relevance paraphrasing on 4 diverse datasets, as well as 4 LLMs of different sizes (GPT-3.5\textsubscript{Turbo}, Llama-2\textsubscript{13B}, Mistral\textsubscript{7B}, and Dolly-v2\textsubscript{7B}). Our results indicate that LLMs are not consistent summarizers for the minimally perturbed articles, necessitating further improvements.  

\end{abstract}

\section{Introduction}
\vspace{-0.1cm}
% Large Language Models (LLMs) have achieved tremendous success at a number of natural language tasks such as question answering \citep{robinson2022leveraging}, computer program generation \citep{vaithilingam2022expectation}, and text summarization \citep{zhang2023benchmarking}, among others. 

Large Language Models (LLMs) have made remarkable progress in generating \textit{abstractive} summaries from input articles that are comparable to summaries written by humans \citep{zhang2023benchmarking}. However, while \textit{best-case} performance of LLMs at zero-shot summarization is clearly superlative to other neural models, relatively little is known about the \textit{robustness} of their performance at this task.

\looseness-1Previous work on LLM robustness has primarily investigated generalizability on discriminative tasks \cite{wang2023causal, wang2022should, wang2023robust, zhou2023context}. One aspect of these tasks is \textit{adversarial robustness} where adversarial prompts meant to induce unsafe behavior are evaluated \citep{zhu2023promptbench, wang2021adversarial}. However, we investigate how robust the generative task of abstractive summarization is when the input article is altered via semantic-preserving perturbations. Similarly, a number of adversarial attacks have been proposed for LLMs for various threat models \citep{jones2023automatically, zou2023universal} based on manual engineering or prompt optimization. However, our goal in this work differs conceptually from an adversarial attack-- we aim to measure \textit{general} robustness performance using a novel paraphrasing strategy which does not have knowledge of the target LLM being used. In contrast, adversarial attacks seek to induce \textit{worst-case} LLM performance by crafting adversarial inputs specific to the model. 

% Note that these attacks target the instruction following capabilities of LLMs, and summarization-specific attacks have not yet been proposed.

%\vspace{-1.5mm}
\begin{figure*}[htb!]
  \centering
  \includegraphics[width=0.8\textwidth]{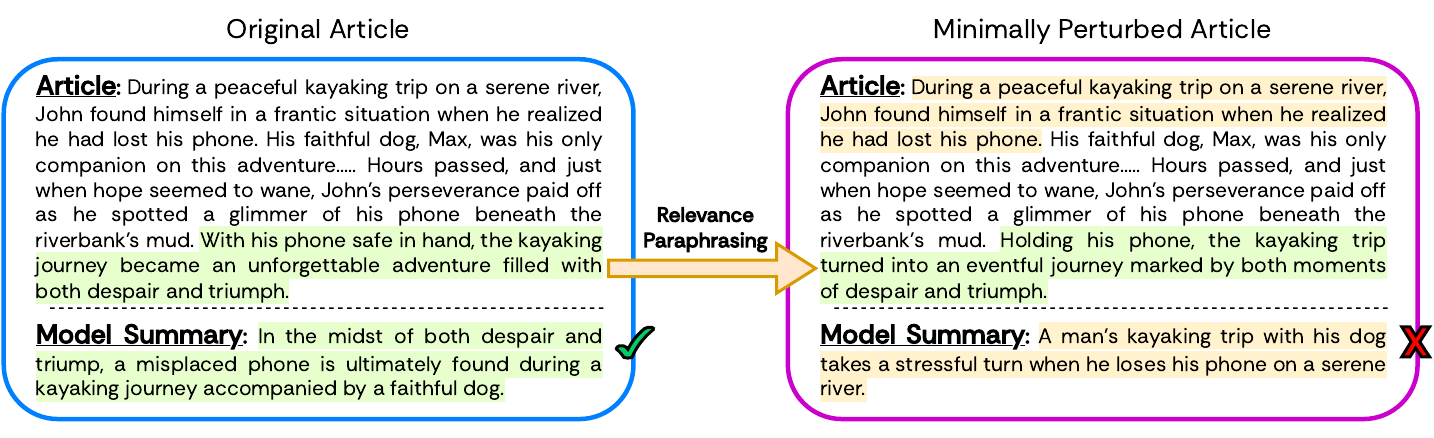}%\vspace{-3mm} %0.95
  \caption{An example showcasing \textit{relevance paraphrasing}. When sentences \textit{relevant} to generating the summary are \textit{paraphrased} to create a minimally perturbed article, we find that summarizaton performance drops as the model uses other sentences instead to craft the summary, leading to a loss of salient information.}\label{fig:blurb}\vspace{-5mm}
\end{figure*}

\looseness-1Other works \citep{ye2023assessing, ko2023robustness} have raised concerns of variability in existing LLM benchmarks and an overall lack of performance credibility (for instance, due to known issues of test set leakage into training data) to measure robustness by proposing novel \textit{evaluation methods}. %There are also a number of position papers \citep{vstefanik2022methods} and surveys \citep{chang2023survey} on robustness in LLMs, but none

\looseness-1To our best knowledge, none of these prior works have explored the robustness of LLM performance at the generative task of \textit{abstractive summarization}. 

In this work, we aim to bridge this gap by proposing a novel method for analyzing the robustness of LLM summarization. For learning tasks, \textit{robustness} has generally been defined \citep{carlini2017towards} as the \textit{change in the magnitude of model performance upon minimally perturbing the input space}. Based on this definition, we formulate and seek to answer the following research question in this work: \textit{how does LLM abstractive summarization performance vary with minimal perturbations of the input articles to be summarized?} 

To make progress towards this goal of quantitatively assessing LLM robustness at summarization, we propose a novel strategy named \textit{relevance paraphrasing} for minimally perturbing the input space of articles. Relevance paraphrasing involves identifying which \textit{relevant} sentences from the input article contribute most to generating an ideal gold summary. Then these sentences are \textit{paraphrased} in the article so that they retain semantic meaning to the original version but are phrased differently. This gives us a semantically equivalent version of the input set of articles as only a few sentences are paraphrased.\footnote{Additional experiments on paraphrasing non-relevant sentences and paraphrasing a larger proportion of the input article are presented in Appendix ~\ref{appendix:non-relevant} and Appendix ~\ref{appendix:larger-proportion}. } Note that paraphrasing is a simple operation that retains close similarity to the original set of articles so if the LLM is a robust summarizer, its performance should not change much for the perturbed input articles. \footnote{There are other methods of perturbing inputs in a semantically equivalent manner (e.g. lexical substitution \citet{mccarthy2007semeval}). However, we chose relevance paraphrasing as lexical substitution does not introduce much syntactical variance and the introduced perturbation is not sufficient to test summarization robustness.} Thus, by measuring the change in performance on both the original and perturbed set of input articles, we can assess LLM summarization robustness. An example of \textit{relevance paraphrasing} is shown in Figure \ref{fig:blurb}. 

More importantly, through our analysis of LLM summarization robustness, we wish to draw attention to the need for more work on task-specific robustness analysis of LLMs. As shown in our results in subsequent sections, LLMs tend to exhibit lower performance across a number of different evaluation metrics (such as ROUGE \citep{lin2004rouge} and BertScore \citep{bertscore}) for the perturbed input articles obtained using relevance paraphrasing. We find that post relevance paraphrasing, LLMs select different input article sentences to craft the output summary, losing salient information in the process. This trend is consistently observed across LLMs of different sizes and model parameters\footnote{We study GPT-3.5\textsubscript{Turbo} \citep{gpt}, Llama-2\textsubscript{13B} \citep{llama2}, Dolly-v2\textsubscript{7B} \citep{dolly}, and Mistral\textsubscript{7B-Instruct-v0.1} \citep{mistral} in all experiments.} as well as multiple datasets. Our results hence indicate that LLMs are not consistent summarizers, and necessitate further improvements to ensure more consistent summarization performance.

\vspace{-0.1cm}

\section{Measuring Robustness Via Relevance Paraphrasing}
\vspace{-0.1cm}
\subsection{Zero-Shot Summarization}

A zero-shot abstractive summarization model $\mathcal{M}$ takes as input $X$ is a set of articles. Each article $x \in X$ has a variable number of sentences. The model $\mathcal{M}$ then takes in as input the set of articles in the set $X$ and outputs a set of summaries, i.e., $\mathcal{M}(X) = S^{\mathcal{M}}$ where $S^{\mathcal{M}}$ is the set of model generated summaries. Traditionally, the model is evaluated by comparing the generated summaries ($S^{\mathcal{M}}$) with \textit{gold standard} summaries written by human experts (denoted as $S^G$) using evaluation metrics such as ROUGE \citep{lin2004rouge} and BertScore \citep{bertscore}. 

\subsection{Relevance Paraphrasing}

Let an article be denoted as $x \in X$ and its corresponding gold summary is $s \in S^{G}$. Similar to previous work in abstractive summarization \citep{reddit, zhao2022narrasum}, we assume a proxy mapping $\psi$ that takes in a (gold) summary sentence $s_i \in s$ and returns a sentence $x_j \in x$ in the article that contributed most to that summary sentence.\footnote{We can also return more than one sentence in this framework as described in Appendix~\ref{appendix:topN}.} Any similarity function can be employed as a useful approximation for such a function $\psi$ but in this paper we utilize TF-IDF vector similarities due to computational efficiency and overall accuracy.\footnote{Note that other metrics may also be considered. We experimented with ROUGE-1 as an alternative and found no significant differences in the results as shown in Appendix~\ref{appendix:psi}.} Also let us assume that we have a paraphrasing model $\theta$ that takes in as input a sentence and returns a paraphrased version which retains semantic similarity but is phrased differently.\footnote{Such a model $\theta$ could be a simple strategy such as \textit{active-to-passive}, \textit{formal-to-casual}, or a neural model such as an LLM being used for paraphrasing.} In this paper, we use Llama-2\textsubscript{13B}\footnote{\looseness-1We use the instruction-tuned Llama-2 variant throughout.} for this purpose.

The \textit{relevance paraphrasing} process is presented as Algorithm~\ref{alg1}. Here, we wish to uncover how robust LLMs are at the task of abstractive summarization. In particular, the process works as follows: we first obtain the gold summary for each input article $x \in X$ as $s \in S^{G}$. Next, we use $\psi$ to obtain a set of article sentences corresponding to each summary sentence in $s$. Analytically, using $\psi$ for each article-summary pair $(x,s)$, let us maintain a set of indices $I_x = \{j | x_j = \psi(s_i), \forall s_i \in s\}$ which is essentially a set of all the article sentence indices that contributed most to the gold summary.

\vspace{-1.5mm}
\begin{algorithm}[H]
\fontsize{8}{8}\selectfont
\caption{: Relevance Paraphrasing}\label{alg1}
 \begin{algorithmic}[1]
 \STATE \textbf{Input:} LLM $\mathcal{M}$, Dataset $T = (X, S^G)$, mapping function $\psi$, paraphrasing model $\theta$, evaluation metric $\mathcal{E}$.
 \STATE \textbf{initialize} $X' = \emptyset$
 \FOR{\textbf{each} $s \in S^{G}$ \textbf{and} $x \in X$ \textbf{pair}}
 \STATE \textbf{let} $I_x = \{j | x_j = \psi(s_i), \forall s_i \in s\}$.
 \STATE \textbf{obtain} $x'$ by replacing $x_i, \forall i \in I_x$ with $\theta(x_i)$.
 \STATE \textbf{obtain} $X' = X' \cup \{ x'\}$.
 \ENDFOR
 \STATE \textbf{measure} $\mathcal{E}(S^G,\mathcal{M}(X))$ \textbf{and} $\mathcal{E}(S^G, \mathcal{M}(X'))$.
\end{algorithmic}
\end{algorithm}
\vspace{-1.5mm}
Now, our goal is to paraphrase each of these \textit{relevant} sentences for article $x$ (that are important for its summary) using the paraphrasing model. We then replace those sentences in the article with their paraphrased versions.\footnote{We perform an automatic evaluation to test the semantic relevance between the original and paraphrased sentences by calculating the BertScore between them-- CNN: 0.9387, XSum 0.9410, News: 0.9320 and Reddit: 0.9190, indicating high semantic similarity between the sentences. We also provide a few qualitative examples in Appendix~\ref{appendix:paraphrasing_examples}, along with other paraphrasing models tested in Appendix~\ref{appendix:paraphrasing_models}.} That is, for each of these article sentences $x_i, \forall i \in I_x$ we will now obtain a paraphrased version $x_i'$ using the paraphrasing model $\theta$ and replace each $x_i$ with paraphrased $x_i'$ to obtain a paraphrased version of the article $x'$. We then repeat this process to obtain the entire set of paraphrased articles as $X'$. Now using the difference in obtained model performance we can assess the summarization robustness of LLMs. For instance, if a given evaluation metric $\mathcal{E}$ (such as BertScore) averaged over all test set summaries worsens (e.g. $\mathcal{E}(S^G,\mathcal{M}(X)) > \mathcal{E}(S^G, \mathcal{M}(X'))$) for the paraphrased set of articles compared to the original versions, we can conclude that the LLM performance is not robust.

% \subsection{Effects of paraphrasing}

\vspace{-0.15cm}

\section{Results}
\vspace{-0.1cm}
% give dataset, model info and parameter values and functions used firsst etc. Point to appendix wherever needed

We now present results for assessing robustness through our proposed relevance paraphrasing strategy. We undertake extensive experiments on 4 LLMs of different sizes: GPT-3.5\textsubscript{Turbo}, Llama-2\textsubscript{13B}, Mistral\textsubscript{7B}, and Dolly-v2\textsubscript{7B}, and 4 diverse real-world datasets: CNN/DM \citep{cnn}, XSum \citep{xsum}, Reddit \citep{reddit}, and News \citep{news}. 
% We use Llama-2\textsubscript{13B} as the paraphrasing model for all experiments. 
Please refer to Appendices \ref{appendix:datasets} and \ref{appendix:models} for detailed information on the datasets and models, respectively.

\begin{table}[!t]
\centering
\caption{Performance change (\%) observed after relevance paraphrasing across datasets/LLMs.}
\label{tab:res}
\resizebox{0.47\textwidth}{!}{%
\begin{tabular}{llllllcl}
\toprule
\multirow{1}{*}{Datasets} & \multirow{1}{*}{Metrics} & \multicolumn{1}{c}{\textbf{Llama-2\textsubscript{13B}}} & \multicolumn{1}{c}{\textbf{GPT-3.5\textsubscript{Turbo}}}  & \multicolumn{1}{c}{\textbf{Dolly-v2\textsubscript{7B}}} & \multicolumn{1}{c}{\textbf{Mistral\textsubscript{7B}}} \\ \cmidrule{3-6}

 &  & \multicolumn{4}{c}{Performance Change (\%)}   \\ \midrule
 & ROUGE-1 & \multicolumn{1}{c}{{\color[HTML]{9A0000}(-)7.354}} & \multicolumn{1}{c}{{\color[HTML]{9A0000}(-)8.750}} & \multicolumn{1}{c}{{\color[HTML]{9A0000} (-)13.77}} & \multicolumn{1}{c}{{\color[HTML]{9A0000}(-)6.814}}  \\
 & ROUGE-2 & \multicolumn{1}{c}{{\color[HTML]{9A0000}(-)21.20}} & \multicolumn{1}{c}{{\color[HTML]{9A0000}(-)23.73}} & \multicolumn{1}{c}{{\color[HTML]{9A0000} (-)31.66}} & \multicolumn{1}{c}{{\color[HTML]{9A0000}(-)27.72}}  \\
 & ROUGE-L & \multicolumn{1}{c}{{\color[HTML]{9A0000}(-)9.431}} & \multicolumn{1}{c}{{\color[HTML]{9A0000}(-)13.54}} & \multicolumn{1}{c}{{\color[HTML]{9A0000} (-)15.70}} & \multicolumn{1}{c}{{\color[HTML]{9A0000}(-)11.99}} &  \\
\multirow{-4}{*}{\textbf{CNN}} & BertScore & \multicolumn{1}{c}{{\color[HTML]{9A0000}(-)0.311}} & \multicolumn{1}{c}{{\color[HTML]{9A0000}(-)0.689}} & \multicolumn{1}{c}{{\color[HTML]{9A0000} (-)5.754}} & \multicolumn{1}{c}{{\color[HTML]{9A0000}(-)0.522}}  \\ \midrule

 & ROUGE-1 & \multicolumn{1}{c}{{\color[HTML]{9A0000}(-)2.837}} & \multicolumn{1}{c}{\color[HTML]{036400} (+)16.19} & \multicolumn{1}{c}{{\color[HTML]{036400} (+)0.680}} & \multicolumn{1}{c}{{\color[HTML]{9A0000}(-)3.680}}  \\
 & ROUGE-2 & \multicolumn{1}{c}{{\color[HTML]{9A0000}(-)8.077}} & \multicolumn{1}{c}{\color[HTML]{036400} (+)12.99} & \multicolumn{1}{c}{{\color[HTML]{9A0000} (-)3.607}} & \multicolumn{1}{c}{{\color[HTML]{9A0000}(-)13.91}}  \\
 & ROUGE-L & \multicolumn{1}{c}{{\color[HTML]{9A0000}(-)3.764}} & \multicolumn{1}{c}{\color[HTML]{036400} (+)11.41} & \multicolumn{1}{c}{{\color[HTML]{036400} (+)1.465}} & \multicolumn{1}{c}{{\color[HTML]{9A0000}(-)3.649}}  \\
\multirow{-4}{*}{\textbf{XSum}} & BertScore & \multicolumn{1}{c}{{\color[HTML]{9A0000}(-)0.092}} & \multicolumn{1}{c}{\color[HTML]{036400} (+)0.321} & \multicolumn{1}{c}{{\color[HTML]{9A0000} (-)0.524}} & \multicolumn{1}{c}{\color[HTML]{036400} (+)0.047}  \\ \midrule

 & ROUGE-1 & \multicolumn{1}{c}{{\color[HTML]{9A0000}(-)10.90}} & \multicolumn{1}{c}{{\color[HTML]{9A0000}(-)15.41}} & \multicolumn{1}{c}{{\color[HTML]{9A0000} (-)39.60}} & \multicolumn{1}{c}{{\color[HTML]{9A0000}(-)7.457}}  \\
 & ROUGE-2 & \multicolumn{1}{c}{{\color[HTML]{9A0000}(-)28.43}} & \multicolumn{1}{c}{{\color[HTML]{9A0000}(-)36.96}} & \multicolumn{1}{c}{{\color[HTML]{9A0000} (-)50.30}} & \multicolumn{1}{c}{{\color[HTML]{9A0000}(-)19.43}}  \\
 & ROUGE-L & \multicolumn{1}{c}{{\color[HTML]{9A0000}(-)13.15}} & \multicolumn{1}{c}{{\color[HTML]{9A0000}(-)17.00}} & \multicolumn{1}{c}{{\color[HTML]{9A0000} (-)41.79}} & \multicolumn{1}{c}{{\color[HTML]{9A0000}(-)10.65}}  \\
\multirow{-4}{*}{\textbf{News}} & BertScore & \multicolumn{1}{c}{{\color[HTML]{9A0000}(-)0.080}} & \multicolumn{1}{c}{{\color[HTML]{9A0000}(-)0.707}} & \multicolumn{1}{c}{{\color[HTML]{9A0000} (-)7.083}} & \multicolumn{1}{c}{\color[HTML]{036400}(+)0.528}  \\ \midrule

 & ROUGE-1 & \multicolumn{1}{c}{\color[HTML]{9A0000} (-)3.158} & \multicolumn{1}{c}{\color[HTML]{9A0000} (-)6.600} & \multicolumn{1}{c}{{\color[HTML]{9A0000} (-)21.85}} & \multicolumn{1}{c}{\color[HTML]{9A0000} (-)2.974}  \\
 & ROUGE-2 & \multicolumn{1}{c}{\color[HTML]{9A0000} (-)13.10} & \multicolumn{1}{c}{\color[HTML]{9A0000} (-)24.13} & \multicolumn{1}{c}{{\color[HTML]{9A0000} (-)13.20}} & \multicolumn{1}{c}{\color[HTML]{9A0000} (-)13.89}  \\
 & ROUGE-L & \multicolumn{1}{c}{\color[HTML]{9A0000} (-)3.529} & \multicolumn{1}{c}{\color[HTML]{9A0000} (-)7.646} & \multicolumn{1}{c}{{\color[HTML]{9A0000} (-)27.64}} & \multicolumn{1}{c}{\color[HTML]{9A0000} (-)1.700}  \\
\multirow{-4}{*}{\textbf{Reddit}} & BertScore & \multicolumn{1}{c}{\color[HTML]{9A0000} (-)0.070} & \multicolumn{1}{c}{\color[HTML]{9A0000} (-)0.750} & \multicolumn{1}{c}{{\color[HTML]{9A0000} (-)18.84}} & \multicolumn{1}{c}{\color[HTML]{036400} (+)2.104}  \\ \bottomrule
\end{tabular}%
}\vspace{-2mm}
\end{table}
\vspace{-0.1cm}
\subsection{LLMs Are Not Consistent Summarizers}
\vspace{-0.1cm}
\looseness-1We present the relative performance change\footnote{That is, $(new - old)/old * 100 \%$.} for the original LLM summary and the one obtained after relevance paraphrasing in Table \ref{tab:res}. We evaluate over 4 holistic summarization metrics: ROUGE-1/2/L and BertScore. We provide the specific original/paraphrased performance bar charts, which further elaborate these trends, in Appendix~\ref{appendix:addn_results}. We also provide results for LLM based evaluation metrics in Appendix~\ref{appendix:geval}, using NLI as an evaluation metric in Appendix ~\ref{appendix:NLI}, when the temperature parameter set to 0 in 
Appendix \ref{appendix:temp0}, traditional summarization models (BART and Pegasus) in \ref{appendix:BART} and results for successive prompting in \ref{appendix:successive}.

% ROUGE-2/L metrics in Figure \ref{fig:res} and defer ones for ROUGE-1 and BertScore showcasing similar trends to 

\begin{figure}[t]
  \centering
  \includegraphics[width=0.45\textwidth]{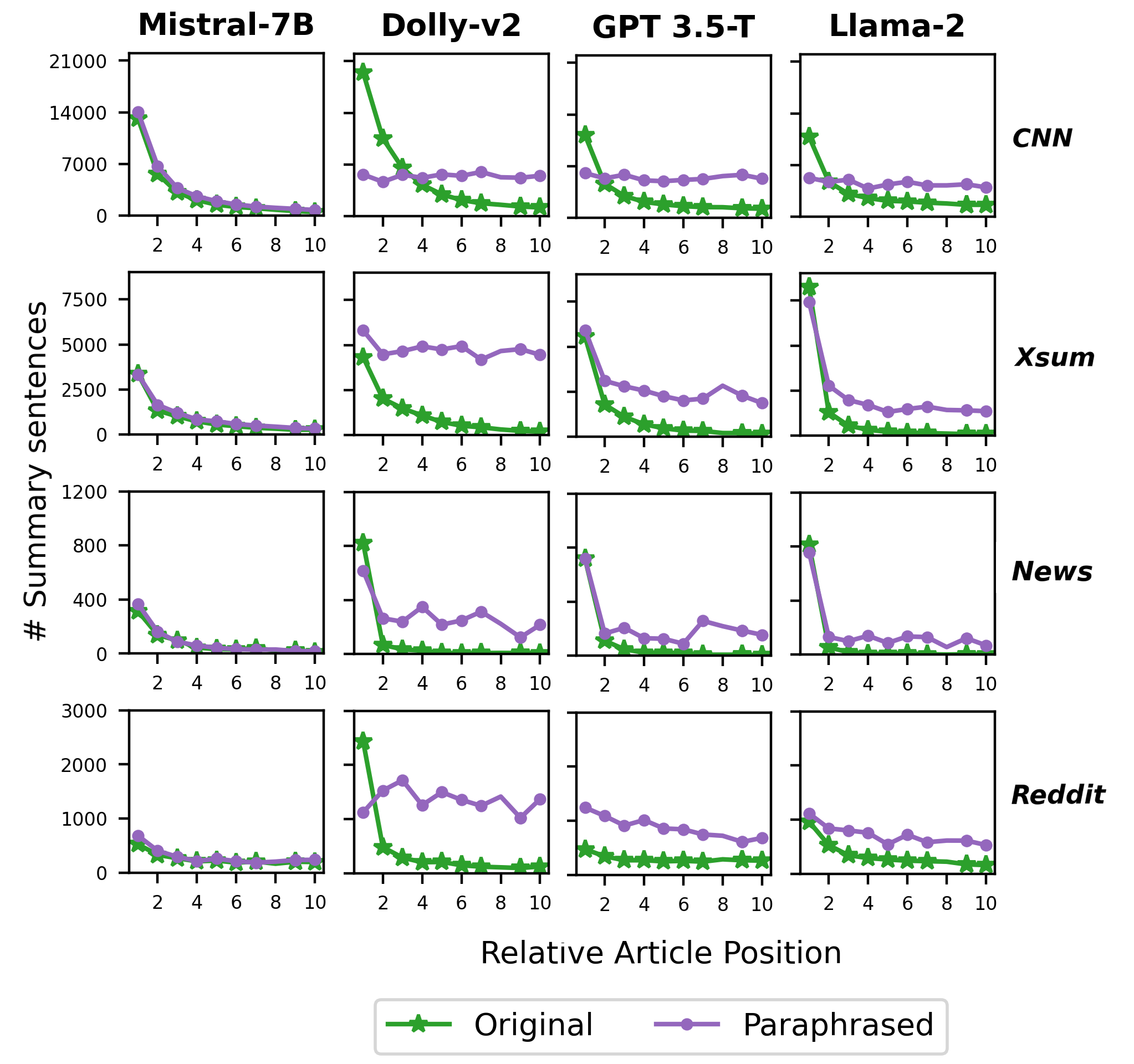}%\vspace{-4mm} %0.65
  \caption{Paraphrasing results in different summaries.}
  \label{fig:pos}\vspace{-4mm}
\end{figure}

\looseness-1Through these results it can be observed that summarization performance drops significantly after relevance paraphrasing for all LLMs. The largest drops observed are for CNN/DM and News-- up to 50\% on ROUGE-2 for Dolly-v2\textsubscript{7B}. Dolly-v2\textsubscript{7B} is the most affected by relevance paraphrasing, with significant drops in performance over all datasets. Even GPT-3.5\textsubscript{Turbo} has performance degradation on the minimally perturbed articles, while Mistral\textsubscript{7B} demonstrates the most robust performance overall. As an exception, GPT-3.5\textsubscript{Turbo} attains large gains in all evaluation metrics after relevance paraphrasing for the XSum dataset. In a few other cases, such as for Mistral (BertScore) and Dolly-v2 (ROUGE), performance has improved post relevance paraphrasing, but only in marginal amounts. These results show that \textit{LLMs are not consistent summarizers, and more improvements need to be made to ensure consistency in outputs}.
\vspace{-0.1cm}

\subsection{Relevance Paraphrasing Leads to Mostly Different LLM Generated Summaries}
%\vspace{-0.1cm}
\looseness-1We explore how LLM summarization selection decisions change as a function of relevance paraphrasing. Using our proxy mapping $\psi$ we can observe the distribution of which input article sentences contributed information to which model summary sentence. We then observe these trends pre and post relevance paraphrasing. These results are shown in Figure \ref{fig:pos}, and it can be seen that LLMs start utilizing different sentences to generate the summary on the paraphrased input article. While this selection issue is somewhat lesser for Mistral\textsubscript{7B}, in general, it poses to be a major problem for all other LLMs. These results further strengthen the finding that LLMs are not consistent summarizers, as \textit{a minor perturbation in the input space leads to significant changes in the output}. 
%\vspace{-0.25cm}

\subsection{Human Evaluation of Summaries}

\looseness-1To further strengthen our results, we conduct \textit{human evaluation} of different article-summary pairs, along the methodological lines of \citet{zhang2024benchmarking,fabbri2021summeval}. We recruit 22 unpaid student annotators, where each annotator was given up to 30 article-summary pair triplets (gold, original and paraphrased). All 4 LLMs were evaluated on the \textit{XSum} and \textit{Reddit} datasets. Each dataset-model pair was annotated by 3 annotators. The results are provided in Appendix~\ref{appendix:HumanEval}.

As can be observed, out of the 8 dataset-model pairs, paraphrased summaries were rated as the best only once (1/8 or 12.5\% of the time). In contrast, original summaries were preferred 34.5\% of the time (3/8) and gold summaries were rated the best 50\% (4/8) of the time. It can be seen that relevance paraphrasing leads to summaries that are not as highly preferred (by humans) as the original (non-paraphrased input) and gold summaries.

\section{Related Works}
\looseness-1LLM robustness has largely been studied in the context of adversarial robustness, where a malicious adversary seeks to execute unsafe model behavior by \textit{automatedly} \citep{zou2023universal, wang2023adversarial, zhu2023autodan} or \textit{manually} optimizing \citep{wei2023jailbroken, perez2022ignore, rao2023tricking} input prompts. Complementary to these efforts, benchmarks have also been proposed to evaluate adversarial robustness of LLMs \citep{zhu2023promptbench, wang2021adversarial}. It is important to note that our work contrasts with research on adversarial robustness of LLMs both conceptually and in terms of motivation. Instead of generating worst-case model specific adversarial prompts, we employ model agnostic relevance paraphrasing to characterize robustness of LLMs at the summarization task.
% To do this we provide semantically equivalent but minimally perturbed input articles to the model. 
Complementarily, recent work has also shown that LLM based abstractive summarization suffers from position bias, further demonstrating their brittleness at the summarization task \cite{chhabra2024revisiting}. %More recently, defense approaches against these attacks have also been proposed \citep{robey2023smoothllm}. 

 To our best knowledge, while a number of works have studied the summarization capabilities of LLMs \citep{tam2023evaluating, zhang2023benchmarking, shen2023large}, none of these have analyzed the robustness of LLMs at the summarization task, which we seek to assess through our work. \footnote{Additional related work for non-LLM neural abstractive
summarization and LLM robustness on tasks other than summarization are presented in Appendix \ref{appendix:relatedwork}.}

%\vspace{-0.15cm}
\section{Conclusion} %going beyond supervised fine-tuning and discuss possible solutions to the problem
%\vspace{-0.15cm}
We propose \textit{relevance paraphrasing} to enable the robustness analysis of LLMs as abstractive summarizers. We find that LLMs are not consistent summarizers, and they begin to use different article sentences to generate summaries for paraphrased articles. Our results indicate that LLMs need further improvements to ensure robustness. 

% By exposing these robustness issues, we believe future work can extend our efforts by proposing \textit{rectification} strategies employed in the instruction finetuning (RLHF) stage\footnote{As sentences can be paraphrased in multiple ways, doing this in the supervised finetuning stage might be intractable.} that resolve these concerns.

\section*{Acknowledgments}
Hadi Askari was supported by the NSF Grant ITE 2333736. Muhao Chen was supported by the DARPA FoundSci Grant HR00112490370, the NSF of the United States Grant ITE 2333736, an Amazon Research Award, and an Amazon Trusted AI Prize. Anshuman Chhabra was supported by the USF CSE department faculty startup fund.

% \clearpage
\section*{Limitations}

Our work analyzes the robustness of LLMs as abstractive summarizers across four diverse datasets. Our results from experiments show that LLMs need to be improved to ensure consistency and robustness in summarization performance (such as via rectification strategies). However, our work has a few limitations that we seek to alleviate in future work. First, summarization robustness needs to assessed in the context of long-form documents (medical records and legal documents, for example) where issues of robustness can lead to adverse outcomes. Second, LLM robustness at summarization needs to be analyzed for low-resource languages and domains where robustness of performance will likely be worsened. Finally, for closed-source models such as GPT-3.5\textsubscript{Turbo}, a longitudinal analysis of summarization robustness needs to be undertaken, as model performance can change over time.

\section*{Ethics Statement}
Our work on uncovering summarization robustness issues in LLMs is important to further improve these models, and ensure robustness of performance. A lack of consistency in generating abstractive summaries can lead to adverse outcomes in real-world scenarios, and our results shed light on this issue through experiments on 4 diverse datasets and 4 different LLMs. Through our initial preliminary efforts, we hope to galvanize research efforts to make LLMs safer and reliable in practice.

\bibliography{refs}
\bibliographystyle{acl_natbib}

%\clearpage

\appendix

\section*{Appendix}

\section{Detailed Dataset Information}\label{appendix:datasets}

\noindent\textbf{CNN/DM} \citep{cnn}:  The CNN/DM dataset contains 300K news articles written by CNN and Daily Mail employees and journalists. The testing set consists of 11490 articles. The average number of sentences in the articles are 33.37 and on average there are 3.79 sentences per summary.

\noindent\textbf{XSum} \citep{xsum}: The XSum dataset contains over 200K short, one-sentence news summaries collected through online articles from the British Broadcasting Corporation. The testing set consists of 11334 articles. The average number of sentences in the articles are 19.105 and on average summaries contain only 1 sentence.

\noindent\textbf{Reddit} \citep{reddit}: The Reddit dataset consists of 120K Reddit posts where these informal crowd-generated posts constitute the text source, in contrast with existing datasets that use formal documents such as news articles as source. We used an 80-20\% train-test split to obtain 4214 articles in the test set. The average number of sentences per article is 22.019 and there are an average of 1.4276 sentences per summary.

\noindent\textbf{News} \citep{news}: The News dataset was initially created for fake news classification. We used the testing set comprising of 1000 articles. In the summaries, there are an average number of 1.012 sentences over all articles.

\section{Detailed Model Information}\label{appendix:models}

\noindent\textbf{GPT-3.5\textsubscript{Turbo}} \citep{gpt}:
GPT-3.5-turbo is OpenAI's flagship LLM which has been instruction-tuned and optimized for chat purposes. We utilized the model using the OpenAI API\footnote{\scriptsize\url{https://platform.openai.com/docs/models/gpt-3-5}} and experiments were conducted on the November version.

\noindent\textbf{Llama-2\textsubscript{13B}} \citep{llama2}: Meta developed the Llama-2 family of LLMs, a collection of pretrained and fine-tuned generative text models ranging in scale from 7-70 parameters. We use the chat version of the models trained via instruction finetuning. We generated inferences via the PyTorch code provided in the official Github repository: \url{https://github.com/facebookresearch/llama}. We used the instruction tuned version of Llama-2\textsubscript{13B} in all experiments.

\noindent\textbf{Dolly-v2\textsubscript{7B}} \citep{dolly}: Dolly is a 6.9 billion parameter causal language model created by Databricks finetuned on a 15K instruction corpus generated by Databricks employees. We used the \textit{databricks/dolly-v2-7b} checkpoint\footnote{\scriptsize\url{https://huggingface.co/databricks/dolly-v2-7b}} from HuggingFace as the summarization model.

\noindent\textbf{Mistral\textsubscript{7B-Instruct-v0.1}} \citep{mistral}: This is the first LLM developed by Mistral AI that is a decoder-based model trained with the following architectural choices: grouped query attention, sliding window attention, and byte-fallback tokenization. Due to these choices, despite Mistral\textsubscript{7B} being a 7B parameter model, it outperforms Llama-2\textsubscript{13B} on a number of evaluation benchmarks.

\section{Llama-2 Prompts for Paraphrasing}\label{appendix:para_prompts} %same here, mention any additional details

To paraphrase the article sentences that corresponded to the dataset summary sentences we leveraged Llama-2. It is important to note that Llama-2 refused to paraphrase 4.93\% of the sentences due to the sentences containing objectionable or problematic language. Therefore we removed all of these articles from both the original and paraphrased datasets before generating the summaries. We now present the prompt used:

\small\textit{You are a helpful assistant that is an expert in paraphrasing sentences.
    Paraphrase the sentence I will provide. Please respond with just the paraphrased version of the sentence. Here is the sentence: {\{Sentence\}}
}\normalsize

Note that \textit{\{Sentence\}} was replaced with the article sentence to obtain the paraphrased sentence. We then replace the original sentence in the article with this version to obtain the minimally perturbed article post relevance paraphrasing.

\section{LLM Prompts for Summarization}\label{appendix:summ_prompts} %mention any capping and other details etc here Hadi

\looseness-1In this section we provide the prompts used to generate both original and paraphrased summaries for each LLM and each dataset. The number of sentences prompted per dataset is equal to the nearest integer of the average number of sentences in the corresponding gold summaries. The prompts were improved iteratively and tailored to each LLM to ensure the most reliable prompt following. However, sometimes the models did not follow the prompt specifications exactly and would generate more summary sentences than required for that dataset. For e.g. Llama-2 followed the prompt exactly 45.99\% while generating the original summaries. Hence, for fair comparison between original and paraphrased summaries we uniformly sampled the number of sentences required from the generated output. We now provide prompts below: 

\subsection{Prompts for GPT-3.5\textsubscript{Turbo}}\label{appendix:gpt_prompts}
\noindent\textbf{\textit{XSum}}: \small\textit{For the following article: \{Article\}. Return a summary comprising of 1 sentence. With each sentence in a numbered list format.\\For example:\\1. First sentence}\normalsize

\noindent\textbf{\textit{CNN/DM}}: \small\textit{For the following article: \{Article\}. Return a summary comprising of 3 sentences. Write each sentence in a dash bulleted format. \\For example:\\1. First sentence\\2. Second sentence\\3. Third sentence}\normalsize

\noindent\textbf{\textit{Reddit}}: \small\textit{For the following article: \{Article\}. Return a summary comprising of 1 sentence. With each sentence in a numbered list format.\\For example:\\1. First sentence}\normalsize

\noindent\textbf{\textit{News}}: \small\textit{For the following article: \{Article\}. Return a summary comprising of 1 sentence. With each sentence in a numbered list format.\\For example:\\1. First sentence}\normalsize

\subsection{Prompts for Llama-2\textsubscript{13B}}\label{appendix:llama_prompts}
\noindent\textbf{\textit{XSum}}: \small\textit{For the following article: \{Article\}. Return a summary comprising of 1 sentence. With each sentence in a numbered list format.\\For example:\\1. First sentence}\normalsize

\noindent\textbf{\textit{CNN/DM}}: \small\textit{For the following article: \{Article\}. Return a summary comprising of 3 sentences. With each sentence in a numbered list format.\\For example:\\1. First sentence\\2. Second sentence\\3. Third sentence}\normalsize

\noindent\textbf{\textit{Reddit}}: \small\textit{For the following article: \{Article\}. Return a summary comprising of 1 sentence. With each sentence in a numbered list format.\\For example:\\1. First sentence}\normalsize

\noindent\textbf{\textit{News}}: \small\textit{For the following article: \{Article\}. Return a summary comprising of 1 sentence. With each sentence in a numbered list format.\\For example:\\1. First sentence}\normalsize

\subsection{Prompts for Dolly-v2\textsubscript{7B}}\label{appendix:dolly_prompts}
\noindent\textbf{\textit{XSum}}: \small\textit{Generate a 1 sentence summary for the given article. Article: {\{Article\}}.}\normalsize

\noindent\textbf{\textit{CNN/DM}}: \small\textit{Generate a 3 sentence summary for the given article. Article: {\{Article\}}. }\normalsize

\noindent\textbf{\textit{Reddit}}: \small\textit{Generate a 1 sentence summary for the given article. Article: {\{Article\}}. }\normalsize

\noindent\textbf{\textit{News}}: \small\textit{Generate a 1 sentence summary for the given article. Article: {\{Article\}}. }\normalsize

\subsection{Prompts for Mistral\textsubscript{7B}}\label{appendix:mistral_prompts}
\noindent\textbf{\textit{XSum}}: \small\textit{For the following article: \{Article\}. Return a summary comprising of 1 sentence. With each sentence in a numbered list format.\\For example:\\1. First sentence}\normalsize

\noindent\textbf{\textit{CNN/DM}}: \small\textit{For the following article: \{Article\}. Return a summary comprising of 3 sentences. With each sentence in a numbered list format.\\For example:\\1. First sentence\\2. Second sentence\\3. Third sentence}\normalsize

\noindent\textbf{\textit{Reddit}}: \small\textit{For the following article: \{Article\}. Return a summary comprising of 1 sentence. With each sentence in a numbered list format.\\For example:\\1. First sentence}\normalsize

\noindent\textbf{\textit{News}}: \small\textit{For the following article: \{Article\}. Return a summary comprising of 1 sentence. With each sentence in a numbered list format.\\For example:\\1. First sentence}\normalsize

Note that \textit{\{Article\}} in each prompt should be replaced by the article to be summarized.

\section{Additional Results on Robustness of LLM Summarization Performance}\label{appendix:addn_results}

\begin{figure}[!htb]
  \centering
  \includegraphics[width=0.48\textwidth]{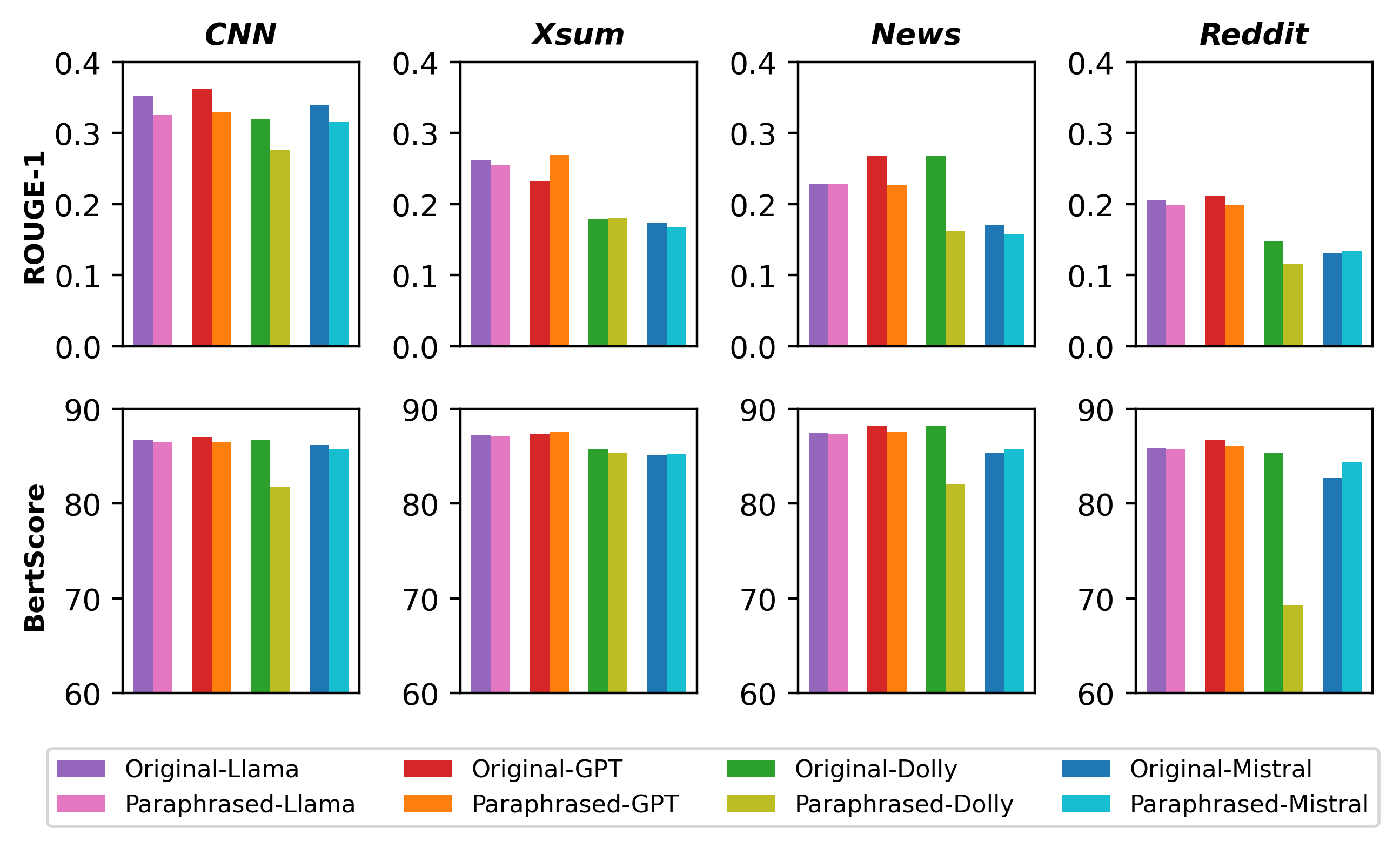}%\vspace{-4mm} %0.65
  \caption{Summarization performance evaluation using ROUGE-1 and BertScore metrics post relevance paraphrasing.}
  \label{fig:appendix_res}\vspace{-4mm}
\end{figure}

\vspace{-1mm}

\begin{figure}[!htb]
  \centering
  \includegraphics[width=0.45\textwidth]{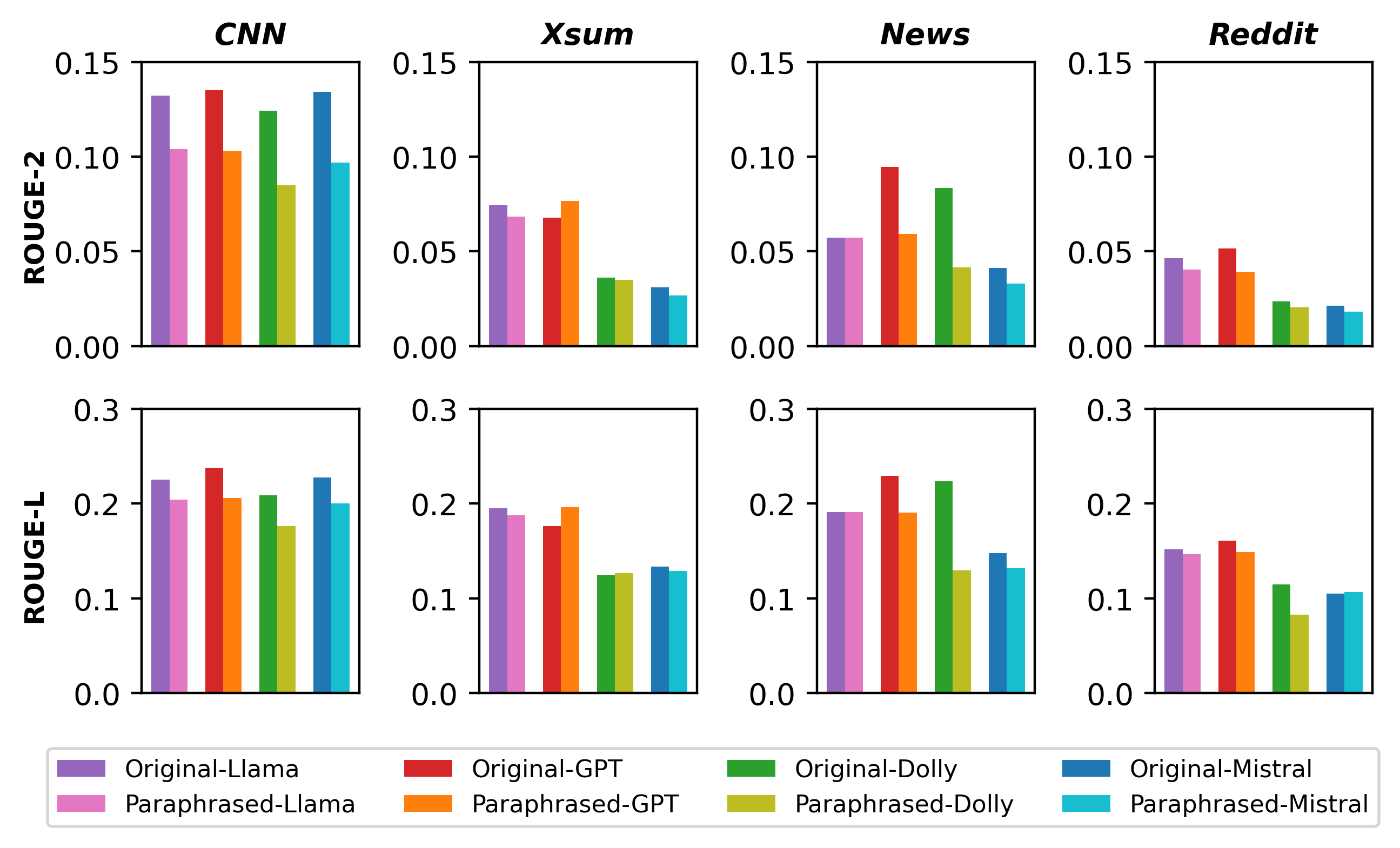}%\vspace{-4mm} %0.65
  \caption{Evaluating summarization performance using ROUGE-2/L on original and paraphrased articles.}
  \label{fig:res}\vspace{-4mm}
\end{figure}

We present results for the BertScore and ROUGE evaluation metrics in Figure \ref{fig:appendix_res} and Figure \ref{fig:res}. It can be seen that for these metrics as well, performance drops consistently across all LLMs post relevance paraphrasing.

\section{Temperature 0 Experiments}\label{appendix:temp0}

\begin{figure}[btp]
  \centering
  \includegraphics[width=0.49\textwidth]{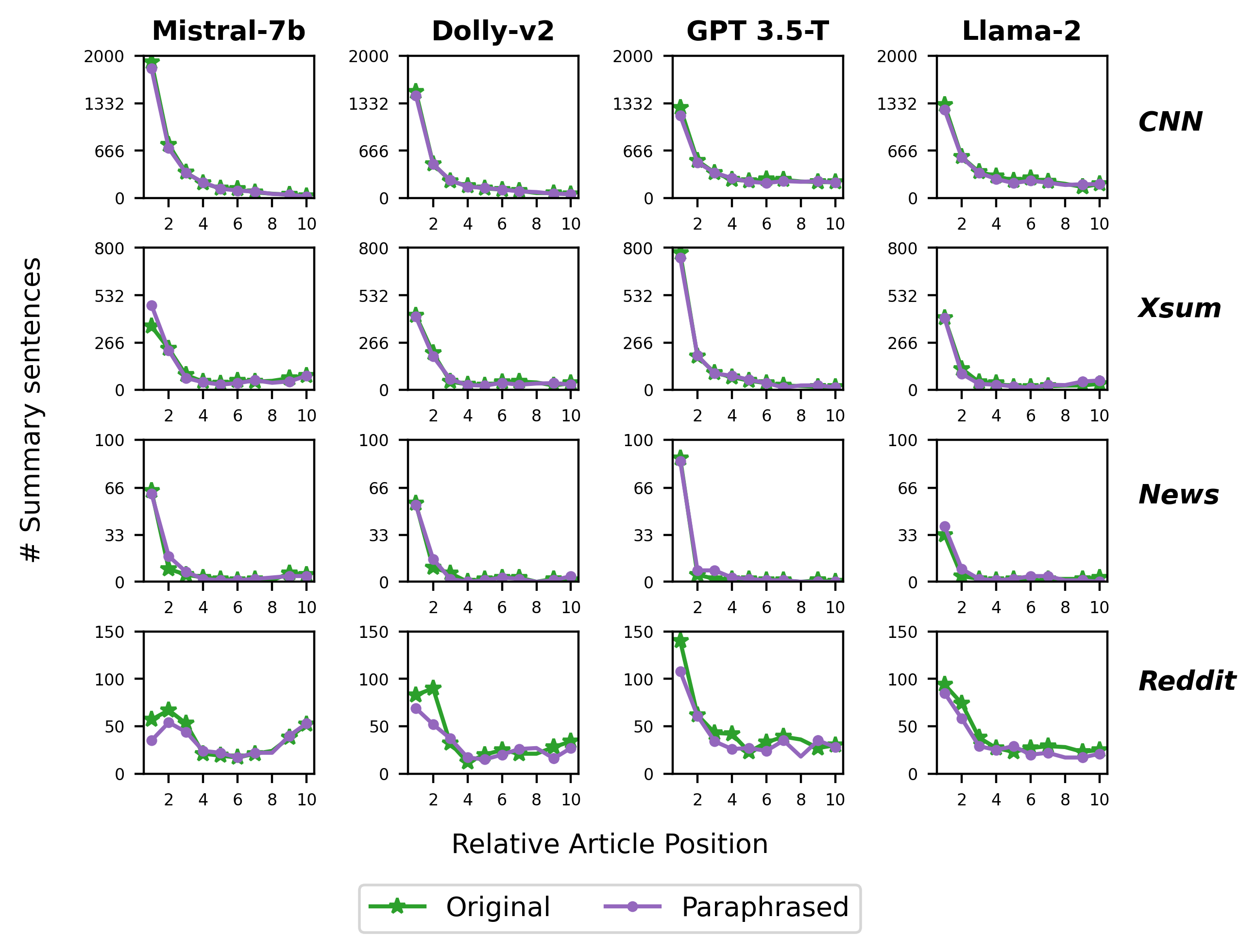}%\vspace{-4mm} %0.65
  \caption{Paraphrasing results in different summaries for Temperature set to 0.}
  \label{fig:appendixpos}\vspace{-4mm}
\end{figure}

We re-run the experiment on a 10\% sample of the datasets while setting the temperature to zero to further investigate the effects of relevance paraphrasing. Note that we did not perform the main experiments with this setting since general and common usage of Large Language models by practitioners has temperature set to non-zero values for stochasticity in outputs. We wanted to assess the robustness in the more realistic general-purpose use-case scenario. Results are presented in Table \ref{tab:temp0} and Figure \ref{fig:appendixpos}. The sentence distribution do not exhibit nearly as much variance as the reference based metrics, this could be a function of the significantly reduced dataset size.

\begin{table}[!htb]
\centering
\caption{Performance change (\%) observed after relevance paraphrasing across datasets/LLMs with Temperature set to 0.}
\label{tab:temp0}
\resizebox{0.49\textwidth}{!}{%
\begin{tabular}{llllllcl}
\toprule
\multirow{1}{*}{Datasets} & \multirow{1}{*}{Metrics} & \multicolumn{1}{c}{\textbf{Llama-2\textsubscript{13B}}} & \multicolumn{1}{c}{\textbf{GPT-3.5\textsubscript{Turbo}}}  & \multicolumn{1}{c}{\textbf{Dolly-v2\textsubscript{7B}}} & \multicolumn{1}{c}{\textbf{Mistral\textsubscript{7B}}} \\ \cmidrule{3-6}

 &  & \multicolumn{4}{c}{Performance Change (\%)}   \\ \midrule
 & ROUGE-1 & \multicolumn{1}{c}{{\color[HTML]{9A0000}(-)7.543}} & \multicolumn{1}{c}{{\color[HTML]{9A0000}(-)5.740}} & \multicolumn{1}{c}{{\color[HTML]{9A0000} (-)10.45}} & \multicolumn{1}{c}{{\color[HTML]{9A0000}(-)10.37}}  \\
 & ROUGE-2 & \multicolumn{1}{c}{{\color[HTML]{9A0000}(-)20.728}} & \multicolumn{1}{c}{{\color[HTML]{9A0000}(-)14.74}} & \multicolumn{1}{c}{{\color[HTML]{9A0000} (-)28.12}} & \multicolumn{1}{c}{{\color[HTML]{9A0000}(-)28.32}}  \\
 & ROUGE-L & \multicolumn{1}{c}{{\color[HTML]{9A0000}(-)9.842}} & \multicolumn{1}{c}{{\color[HTML]{9A0000}(-)7.034}} & \multicolumn{1}{c}{{\color[HTML]{9A0000} (-)13.90}} & \multicolumn{1}{c}{{\color[HTML]{9A0000}(-)14.33}} &  \\
\multirow{-4}{*}{\textbf{CNN}} & BertScore & \multicolumn{1}{c}{{\color[HTML]{9A0000}(-)0.388}} & \multicolumn{1}{c}{{\color[HTML]{9A0000}(-)0.231}} & \multicolumn{1}{c}{{\color[HTML]{9A0000} (-)0.353}} & \multicolumn{1}{c}{{\color[HTML]{9A0000}(-)0.373}}  \\ \midrule

 & ROUGE-1 & \multicolumn{1}{c}{{\color[HTML]{9A0000}(-)5.381}} & \multicolumn{1}{c}{\color[HTML]{9A0000} (-)3.448} & \multicolumn{1}{c}{{\color[HTML]{9A0000} (-)0.637}} & \multicolumn{1}{c}{{\color[HTML]{036400}(+)1.703}}  \\
 & ROUGE-2 & \multicolumn{1}{c}{{\color[HTML]{9A0000}(-)12.405}} & \multicolumn{1}{c}{\color[HTML]{9A0000} (-)8.361} & \multicolumn{1}{c}{{\color[HTML]{9A0000} (-)4.547}} & \multicolumn{1}{c}{{\color[HTML]{036400}(+)1.414}}  \\
 & ROUGE-L & \multicolumn{1}{c}{{\color[HTML]{9A0000}(-)4.377}} & \multicolumn{1}{c}{\color[HTML]{9A0000} (-)3.298} & \multicolumn{1}{c}{{\color[HTML]{9A0000} (-)1.420}} & \multicolumn{1}{c}{{\color[HTML]{036400}(+)1.324}}  \\
\multirow{-4}{*}{\textbf{XSum}} & BertScore & \multicolumn{1}{c}{{\color[HTML]{9A0000}(-)0.249}} & \multicolumn{1}{c}{\color[HTML]{9A0000} (-)0.125} & \multicolumn{1}{c}{{\color[HTML]{9A0000} (-)1.751}} & \multicolumn{1}{c}{\color[HTML]{036400} (+)0.218}  \\ \midrule

 & ROUGE-1 & \multicolumn{1}{c}{{\color[HTML]{036400}(+)7.095}} & \multicolumn{1}{c}{{\color[HTML]{9A0000}(-)2.876}} & \multicolumn{1}{c}{{\color[HTML]{9A0000} (-)19.37}} & \multicolumn{1}{c}{{\color[HTML]{9A0000}(-)11.38}}  \\
 & ROUGE-2 & \multicolumn{1}{c}{{\color[HTML]{9A0000}(-)3.272}} & \multicolumn{1}{c}{{\color[HTML]{9A0000}(-)17.79}} & \multicolumn{1}{c}{{\color[HTML]{9A0000} (-)40.93}} & \multicolumn{1}{c}{{\color[HTML]{9A0000}(-)34.03}}  \\
 & ROUGE-L & \multicolumn{1}{c}{{\color[HTML]{036400}(+)4.312}} & \multicolumn{1}{c}{{\color[HTML]{9A0000}(-)4.335}} & \multicolumn{1}{c}{{\color[HTML]{9A0000} (-)22.17}} & \multicolumn{1}{c}{{\color[HTML]{9A0000}(-)15.59}}  \\
\multirow{-4}{*}{\textbf{News}} & BertScore & \multicolumn{1}{c}{{\color[HTML]{036400}(+)0.436}} & \multicolumn{1}{c}{{\color[HTML]{036400}(+)0.459}} & \multicolumn{1}{c}{{\color[HTML]{9A0000} (-)0.331}} & \multicolumn{1}{c}{\color[HTML]{9A0000}(-)0.103}  \\ \midrule

 & ROUGE-1 & \multicolumn{1}{c}{\color[HTML]{036400} (+)2.174} & \multicolumn{1}{c}{\color[HTML]{9A0000} (-)1.992} & \multicolumn{1}{c}{{\color[HTML]{9A0000} (-)23.87}} & \multicolumn{1}{c}{\color[HTML]{9A0000} (-)6.461}  \\
 & ROUGE-2 & \multicolumn{1}{c}{\color[HTML]{9A0000} (-)1.813} & \multicolumn{1}{c}{\color[HTML]{9A0000} (-)9.880} & \multicolumn{1}{c}{{\color[HTML]{9A0000} (-)41.31}} & \multicolumn{1}{c}{\color[HTML]{9A0000} (-)25.44}  \\
 & ROUGE-L & \multicolumn{1}{c}{\color[HTML]{9A0000} (-)0.704} & \multicolumn{1}{c}{\color[HTML]{9A0000} (-)4.075} & \multicolumn{1}{c}{{\color[HTML]{9A0000} (-)25.82}} & \multicolumn{1}{c}{\color[HTML]{9A0000} (-)6.445}  \\
\multirow{-4}{*}{\textbf{Reddit}} & BertScore & \multicolumn{1}{c}{\color[HTML]{9A0000} (-)0.020} & \multicolumn{1}{c}{\color[HTML]{9A0000} (-)0.102} & \multicolumn{1}{c}{{\color[HTML]{9A0000} (-)22.33}} & \multicolumn{1}{c}{\color[HTML]{9A0000} (-)0.053}  \\ \bottomrule
\end{tabular}%
}\vspace{-2mm}
\end{table}

\section{G-Eval}\label{appendix:geval}
%% Just my notes not official writing
To further evaluate the quality of the summaries we leverage LLM based evaluation, G-Eval \cite{liu2023gpteval}. We used GPT-3.5\textsubscript{Turbo} as the base model and set the temperature to 0. We prompted the LLM-evaluator to re-port back a weighted score from 1 to 5 where 5 is the highest quality. The detailed prompt can be found here Appendix~\ref{appendix:g-evalprompt}.

The results are presented in Table \ref{tab:geval}. The results are inconsistent with the Gold summaries being the highest rated in 9 comparisons, the Original summaries being the highest rated in 3 comparisons and the Paraphrased summaries being the highest rated in 4 comparisons. We find that this lack of consistency in summarization evaluation has been emphasized and observed in past work \cite{zheng2024judging}. The identified issues are as follows: LLM-based evaluators suffer from position bias (prefer the first summary over the second), verbosity bias (prefer longer summaries over shorter ones) and self-enhancement bias (preferring outputs generated by themselves) \cite{zheng2024judging}. These results may suffer from verbosity bias and self-enhancement bias. The GPT-3.5\textsubscript{Turbo} specifically might suffer from self-enhancement bias as the original summaries were generated in September 2023 whereas the paraphrased summaries were generated in December 2023 and OpenAI updates the models regularly. 

% Limitations: Prefer longer even if has factual errors
% Suffers from positional bias, prefers the first summary to the second
% Suffers from self-language bias. Would prefer summaries written in their own style.
% Original summaries were generated by September 2023 version of GPT 3.5 Turbo whereas Paraphrased summaries were generated by the December 2023 version. Reasonable to assume 1) Dec model probably got better 2) Dec model is closer to April 2024 model so it would rate it higher due to self language bias. 

\subsection{G-Eval Prompt:} \label{appendix:g-evalprompt}

\textit{\textbf{Prompt:}}   
\textit{You will be given one summary written for an article. Your task is to rate the summary based on the following criteria:}
    \textit{Output format: PERCENTAGE, PERCENTAGE, PERCENTAGE, PERCENTAGE, PERCENTAGE}

\textit{Evaluation Criteria:
1. Read the news article carefully and identify the main topic and key points.
2. Read the summary and compare it to the news article. Check if the summary covers the main topic and key points of the news article, and if it resents them in a clear and logical order.
3. Rate the summary with 5 percentages, where each one represents how likely the summary is going to get a score from 1 to 5. For example, if you think the summary is 80\% likely to get a score of 5, 10\% likely to get a score of 4, 5\% likely to get a score of 3, 3\% likely to get a score of 2, and 1\% likely to get a score of 1, you should rate the summary as 80, 10, 5, 3, 2.}

\textit{Here is the article: \{Article\}}

\textit{Here is the summary: \{Summary\}}

\begin{table}[!htb]
\centering
\caption{G-Eval (GPT-3.5 Turbo) scores on the three different types of summaries.}
\label{tab:geval}
\resizebox{0.49\textwidth}{!}{%
\begin{tabular}{lccccccc}
\toprule
\multirow{2}{*}{Datasets} & \multirow{2}{*}{Summary} & \multirow{2}{*}{\textbf{Llama-2\textsubscript{13B}}} & \multirow{2}{*}{\textbf{GPT-3.5\textsubscript{Turbo}}}  & \multirow{2}{*}{\textbf{Dolly-v2\textsubscript{7B}}} & \multirow{2}{*}{\textbf{Mistral\textsubscript{7B}}}  \\ 

%\cmidrule{3-6}

 % &  & \multicolumn{4}{c}{Performance Change (\%)}   
 \\ \midrule
 & \multirow{2}{*}{Gold} & \multirow{2}{*}{4.786} & \multirow{2}{*}{4.786} & \multirow{2}{*}{\textbf{4.786}} & \multirow{2}{*}{\textbf{4.786}} \\
 
 & \multirow{2}{*}{Original} & \multirow{2}{*}{\textbf{4.811}} & \multirow{2}{*}{4.783} & \multirow{2}{*}{4.765} & \multirow{2}{*}{4.775}  \\
 & \multirow{2}{*}{Paraphrased} & \multirow{2}{*}{4.809} & \multirow{2}{*}{\textbf{4.814}} & \multirow{2}{*}{4.785} & \multirow{2}{*}{4.751}  \\
\multirow{-4}{*}{\textbf{CNN}} \\

\midrule

 & \multirow{2}{*}{Gold} & \multirow{2}{*}{\textbf{4.798}} & \multirow{2}{*}{\textbf{4.798}} & \multirow{2}{*}{\textbf{4.798}} & \multirow{2}{*}{\textbf{4.798}}  \\
 & \multirow{2}{*}{Original} & \multirow{2}{*}{4.772}& \multirow{2}{*}{4.721} & \multirow{2}{*}{4.797} & \multirow{2}{*}{4.635} 

 \\
 & \multirow{2}{*}{Paraphrased} & \multirow{2}{*}{4.758} & \multirow{2}{*}{4.728} &  \multirow{2}{*}{4.539} & \multirow{2}{*}{4.642} \\
\multirow{-4}{*}{\textbf{Xsum}} \\

\midrule

 & \multirow{2}{*}{Gold} & \multirow{2}{*}{4.753} & \multirow{2}{*}{4.753} & \multirow{2}{*}{4.753} & \multirow{2}{*}{\textbf{4.753}}  \\
 & \multirow{2}{*}{Original} & \multirow{2}{*}{4.799} & \multirow{2}{*}{4.702} & \multirow{2}{*}{\textbf{4.794}} & \multirow{2}{*}{4.648} \\
 & \multirow{2}{*}{Paraphrased} & \multirow{2}{*}{\textbf{4.800}} & \multirow{2}{*}{\textbf{4.806}} & \multirow{2}{*}{4.784}& \multirow{2}{*}{4.643} \\
\multirow{-4}{*}{\textbf{News}} \\

\midrule

 & \multirow{2}{*}{Gold} & \multirow{2}{*}{4.558} & \multirow{2}{*}{4.558} & \multirow{2}{*}{\textbf{4.558}} & \multirow{2}{*}{\textbf{4.558}}  \\
 & \multirow{2}{*}{Original} &  \multirow{2}{*}{\textbf{4.745}} & \multirow{2}{*}{4.713} & \multirow{2}{*}{4.348} & \multirow{2}{*}{4.193}
\\
 & \multirow{2}{*}{Paraphrased} & \multirow{2}{*}{4.733} & \multirow{2}{*}{\textbf{4.782}} & \multirow{2}{*}{4.338} & \multirow{2}{*}{4.208} 
\\
\multirow{-4}{*}{\textbf{Reddit}} \\

\bottomrule
\end{tabular}%
}\vspace{-2mm}
\end{table}

\section{Paraphrasing Examples}\label{appendix:paraphrasing_examples}

To further illustrate the paraphrasing done by Llama-2\textsubscript{13B} we provide a few examples from the CNN dataset:

\begin{itemize}
    \item \textbf{Original:} They were exposed to Ebola in Sierra Leone in March, but none developed the deadly virus.

\item \textbf{Paraphrased:} They encountered Ebola in Sierra Leone in March, yet none of them contracted the fatal infection.

\item \textbf{Original:} The student was identified during an investigation by campus police and the office of student affairs and admitted to placing the noose on the tree early Wednesday, the university said.

\item \textbf{Paraphrased:} The university reported that the student was discovered by campus police and the office of student affairs to have placed a noose on a tree on Wednesday morning, following an investigation.

\item \textbf{Original:} Four days after her apparent death, the dog managed to stagger to a nearby farm, dirt-covered and emaciated, where she was found by a worker who took her to a vet for help.

\item \textbf{Paraphrased:} The dog, who had been presumed dead for four days, miraculously made her way to a nearby farm, badly injured and severely underweight, where she was discovered by a worker who rushed her to a veterinarian for medical attention.
\end{itemize}

Additionally, we also provide a few examples from the Reddit dataset that had the highest drop in BertScore between the original and paraphrased article sentences. 

\begin{itemize}
    \item \textbf{Original:} I accidentally knocked the cupcakes out of her hand, causing them to spill all over her.
    \item \textbf{Paraphrased:} i knocked the cupcakes out of her hand and they spilled  all over her.
    \item \textbf{Original:}  i was just now preparing for bed, brushing my teeth, when i reached for my retainer and found it...gone!
    \item \textbf{Paraphrased:} i was getting ready for bed, brushing my teeth, when I realized that my retainer was nowhere to be found.
    \item \textbf{Original:}  well as soon as i asked him for a raise, he looked up at me, smiled and pulled out a stack of papers from his desk
    \item \textbf{Paraphrased:} 'As soon as I requested a raise, my boss looked up at me with a smile and pulled out a large pile of papers from his desk.
    \item \textbf{Original:} my 5th period isn't the greatest.
    \item \textbf{Paraphrased:} My fifth period is not particularly outstanding.
    \item \textbf{Original:} this was five years ago ( im 18 now ), me and my girlfriend were bored and my parents were not at home.'
    \item \textbf{Paraphrased:} 'About five years ago, when I was 18 years old, my girlfriend and I were feeling bored and our parents were not present at home.
    
\end{itemize}

\section{Paraphrasing Non-Relevant Sentences}\label{appendix:non-relevant}

We generated Paraphrased summaries for the CNN and Reddit datasets on 2000 articles each by randomly paraphrasing the non-relevant sentences. These "non-relevant" sentences were the sentences not selected by our relevance mapping function. We selected an equal amount of non-relevant sentences to paraphrase as we would have selected relevant sentences to paraphrase (for e.g if the summary comprised of 2 sentences, we then paraphrased any two, randomly selected, non-relevant sentences in the input article). We then evaluate and compare with our relevance paraphrasing approach using the Llama-2 LLM. 

\begin{table}[!ht]
\centering
\resizebox{0.45\textwidth}{!}{
\begin{tabular}{ccccc}
\hline
                   & \multicolumn{2}{c}{\textbf{CNN}}    & \multicolumn{2}{c}{\textbf{Reddit}} \\
                   \hline
\textbf{Metrics}   & \textbf{Relevant} & \textbf{Random} & \textbf{Relevant} & \textbf{Random} \\
\hline
ROUGE-1   & {\color[HTML]{9A0000}(-)7.135}           & {\color[HTML]{9A0000}(-)1.625}          & {\color[HTML]{9A0000}(-)3.880}            & {\color[HTML]{9A0000}(-)0.665}          \\

ROUGE-2   & {\color[HTML]{9A0000}(-)19.84}            & {\color[HTML]{9A0000}(-)3.636}          & {\color[HTML]{9A0000}(-)16.17}            & {\color[HTML]{9A0000}(-)2.212}          \\

ROUGE-L   & {\color[HTML]{9A0000}(-)8.498}            & {\color[HTML]{9A0000}(-)1.394}          & {\color[HTML]{9A0000}(-)4.178}            & {\color[HTML]{9A0000}(-)0.179}          \\

BertScore & {\color[HTML]{9A0000}(-)0.238}            & {\color[HTML]{9A0000}(-)0.023}          & {\color[HTML]{9A0000}(-)0.078}            & {\color[HTML]{036400}(+)0.060}   \\
\hline

\end{tabular}}
\end{table}

We observe that paraphrasing random sentences leads to significantly less effect on the output summaries' quality (as measured by evaluation metrics) compared to our relevance paraphrasing approach. For instance, the drop in ROUGE-1 increases from -7.135  (relevance paraphrasing) to -1.625 (non-relevance paraphrasing)  in the CNN dataset and the drop in ROUGE-1 increases from -3.880  (relevance paraphrasing) to -0.665 (non-relevance paraphrasing) in the Reddit dataset. This difference is significant and consistent across the other metrics as well.

\section{Paraphrasing a larger proportion of the article}\label{appendix:larger-proportion}

We paraphrase the top-3 most relevant sentences (as opposed to only the top-1) on a 2000 subsample of the XSum and News datasets and employ the Llama-2 LLM. This experiment thus seeks to increase the proportion of input article sentences that are being paraphrased and observe the robustness of LLM generated summaries.

\begin{table}[!ht]
\centering
\resizebox{0.40\textwidth}{!}{
\begin{tabular}{ccccc}
\hline
                   & \multicolumn{2}{c}{\textbf{XSum}}    & \multicolumn{2}{c}{\textbf{NEWS}} \\
                   \hline
\textbf{Metrics}   & \textbf{Top 1} & \textbf{Top 3} & \textbf{Top 1} & \textbf{Top 3} \\
\hline
ROUGE-1   & {\color[HTML]{9A0000}(-)2.483}           & {\color[HTML]{9A0000}(-)3.091}          & {\color[HTML]{9A0000}(-)10.58}            & {\color[HTML]{9A0000}(-)11.21}          \\

ROUGE-2   & {\color[HTML]{9A0000}(-)5.926}            & {\color[HTML]{9A0000}(-)8.302}           & {\color[HTML]{9A0000}(-)28.32}          & {\color[HTML]{9A0000}(-)33.22}          \\

ROUGE-L   & {\color[HTML]{9A0000}(-)2.644}            & {\color[HTML]{9A0000}(-)3.831}         & {\color[HTML]{9A0000}(-)12.91}           & {\color[HTML]{9A0000}(-)13.64}       \\

BertScore & {\color[HTML]{9A0000}(-)0.050}          & {\color[HTML]{9A0000}(-)0.076}        & {\color[HTML]{9A0000}(-)0.015}            & {\color[HTML]{9A0000}(-)0.022}   \\
\hline

\end{tabular}}
\end{table}

We can observe some performance degradation incurred by paraphrasing more sentences. The BertScore drops from -0.050 to -0.076 in XSum and from -0.015 to -0.022 in News. Hence, we can conclude that relevance paraphrasing with a larger proportion of the input space leads to a further slight decrease in summarization performance, while trading off potential semantic similarity with the original article. We would like to emphasize that the original relevance paraphrasing approach applied minimal perturbation while providing similar evidence of summarization robustness.

\section{Other Paraphrasing Models}\label{appendix:paraphrasing_models}

We investigate other non-LLM paraphrasing models "Pegasus Paraphrase" \cite{pegasusparaphrase} and "chatgpt paraphraser on T5 base" \cite{chatgpt_paraphraser}, since they were popular and well performing on Huggingface. We eventually settled on using Llama-2\textsubscript{13B} as the paraphraser as we saw that the quality of the outputs was significantly better.

To quantify this, we paraphrased 10 sentences each with these 3 models and calculated the BertScore and ROUGE-1 score with the original sentence. 

\begin{table}[!ht]
\centering
\caption{Sampled ROUGE-1 and Bertscore scores on different paraphrasing models.}\label{table:corr}
\resizebox{0.45\textwidth}{!}{%
\begin{tabular}{ccc} 
\hline
\textbf{Model} & \textbf{Bertscore} & \textbf{ROUGE-1} \\  
\hline
\multirow{1}{*}{\textit{\textbf{Pegasus Paraphrase}}} & 94.29 & 80.17 \\
\multirow{1}{*}{\textit{\textbf{chatgpt paraphraser on T5 base}}} & 93.92 & 72.20 \\
\multirow{1}{*}{\textit{\textbf{Llama-2\textsubscript{13B}}}}& 92.16 & 60.64 \\
\hline
\end{tabular}}
\end{table}

Qualitatively, we analyze a few examples to demonstrate:

\begin{itemize}
\item{\textbf{Original:} Zarif left the country in 1977, received his undergraduate degree from San Francisco State University in 1981, his master's in international relations from the University of Denver in 1984 and his doctorate from the University of Denver in 1988.}
\item{\textbf{"Pegasus Paraphrase":} Zarif received his undergraduate degree from San Francisco State University in 1981 and his master's in international relations from the University of Denver in 1984 while he was still in the country.}
\item{\textbf{"chatgpt paraphraser on T5 base":} In 1977, Zarif departed from the country, but in 1981, he received his undergraduate degree from San Francisco State University, a master's degree in international relations from the University of Denver, and a doctorate from the University of Denver in 1988.}
\item{\textbf{Llama-2\textsubscript{13B}}: Zarif departed the country in 1977, later earning his undergraduate degree from San Francisco State University in 1981, his master's in international relations from the University of Denver in 1984, and his doctorate from the same institution in 1988.}
\end{itemize}

Here, "Pegasus Paraphrase" omits the detail that Zarif left the country in 1977 and "chatgpt paraphraser on T5 base" omits the detail of when he received his masters. Llama2-13B, on the other hand, keeps all the pertinent details while paraphrasing as well.

Looking at another example:

\begin{itemize}
    \item \textbf{Original:} Looking spry at 91, Barker handled the first price-guessing game of the show, the classic "Lucky Seven," before turning hosting duties over to Carey, who finished up.

\item \textbf{"Pegasus Paraphrase":} Barker handled the first price-guessing game of the show, the classic "Lucky Seven," before handing hosting duties to Carey, who finished up.

\item \textbf{"chatgpt paraphraser on T5 base":} The first game of the show, the classic "Lucky Seven," was handled by Barker, who then handed over the hosting duties to Carey, who finished up the show looking sappy at 91.

\item \textbf{Llama-2\textsubscript{13B}}: Barker, who appeared youthful at 91, kicked off the first price-guessing game of the show, "Lucky Seven," before passing the baton to Carey to conclude the segment.
\end{itemize}

Here, "Pegasus Paraphrase" again misses the detail that Barker was looking spry at 91 years of age. "chatgpt paraphraser on T5 base" misrepresents the information and paraphrases that Carey was the one who is 91 when actually it was Barker. Finally, Llama2-13B paraphrases the sentence elegantly.

\section{Additional Results for Different $\psi$}\label{appendix:psi}

\begin{figure}[!ht]
  \centering
  \includegraphics[width=0.48\textwidth]{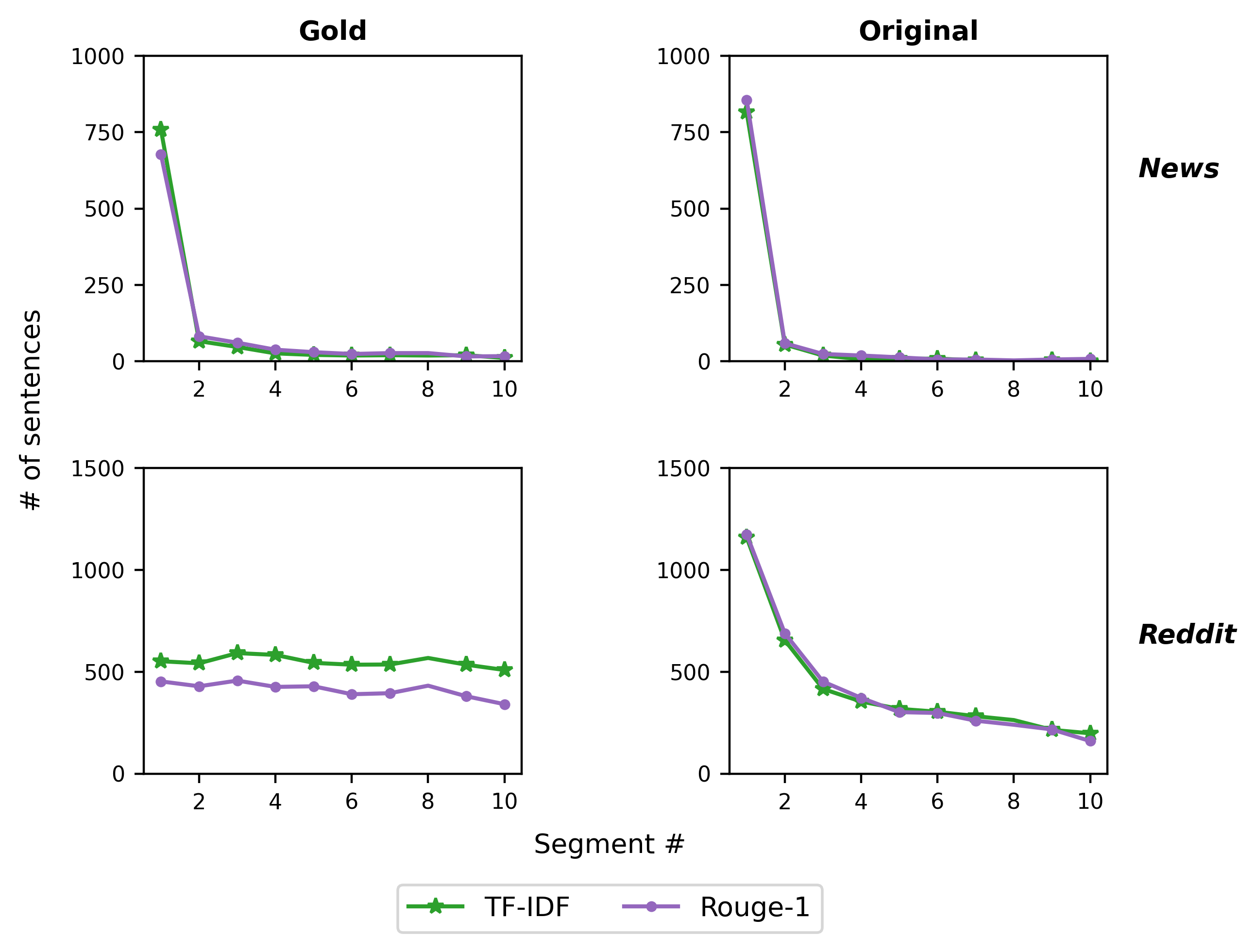}\vspace{-1mm} %0.525
  \caption{Results on \textit{News} and \textit{Reddit} for Llama-2 when $\psi$ is either TF-IDF similarity or ROUGE-1.}
  \label{fig:appendix_rouge}
\end{figure}

\looseness-1For experiments in the main paper, we opt for TF-IDF vector similarities as the choice of the mapping function $\psi$ due to computational efficiency (over computing individual ROUGE scores between summary and article sentences for e.g.). However, it is important to examine whether this choice significantly impacts results, trends, and our findings. In initial experiments with different $\psi$ we concluded that this choice does not affect results. In Figure \ref{fig:appendix_rouge} we provide results that support this by using ROUGE-1 as the metric for $\psi$ on the \textit{Reddit} and \textit{News} datasets for Llama-2 generated summaries. We compare the gold summary and original summary positional distributions for both datasets when $\psi$ is computed using TF-IDF vectors and ROUGE-1. It is clear that the trends and results are the same for both $\psi$.

\section{Mapping Summary Sentences to Top-N relevant Article Sentences}\label{appendix:topN}

Currently, $\psi$ maps back from one summary sentence to one article sentence that contributes the most to that summary sentence. To do this, as $\psi$ measures similarity between sentences, we currently only pick the article sentence with the maximum similarity to the summary sentence. However, since $\psi$ is basically measuring similarity, we can return the top-2 or top-3 matches and undertake the same sentence distribution analysis as in the main results. No specific change is necessary, since our sentence distribution estimation is done in aggregate, via binning. It can be seen that the distributions do change slightly, but overall the trends remain the same. It is beneficial to assess the impact of utilizing multiple article sentences, especially for datasets like \textit{XSum} where the summary is usually just one sentence and discusses facts from multiple article sentences.

\begin{figure}[!ht]
  \centering
  \includegraphics[width=0.49\textwidth]{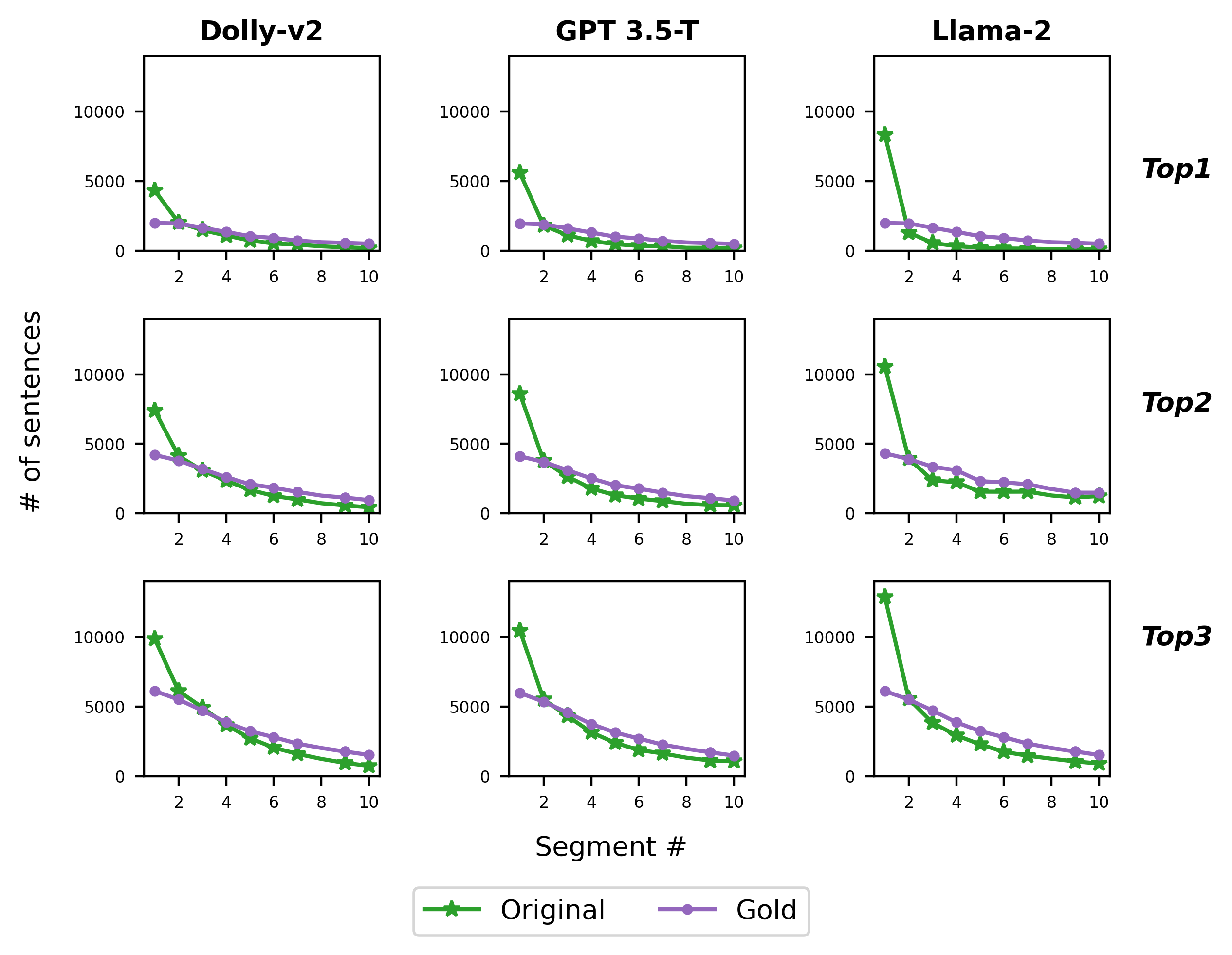}\vspace{-1mm} %0.525
  \caption{Mapping summary sentences to multiple article sentences on \textit{XSum}.}
  \label{fig:appendix_topn}
\end{figure}

\section{Additional Related Work}\label{appendix:relatedwork}

\subsection{LLM Robustness on tasks other than Summarization}

Other work on LLM robustness has proposed evaluation workflows to assess model performance at general instruction following \citep{sun2023evaluating}, program synthesis \citep{shirafuji2023exploring}, sentence classification \citep{ko2023robustness}, reasoning problems \citep{ye2023assessing}, enhancing generalizability via debiasing outputs \cite{wang2023robust, wang2022should, wang2023causal} and mitigating prediction shortcuts \cite{zhou2023context}.

\subsection{Summarization Robustness for non-LLMs}

Prior work has also studied the robustness of non-LLM neural abstractive summarization models. However, these works have traditionally focused on summary factuality/faithfulness as a proxy for measuring robustness \cite{song2021towards,fan2018robust,chen2022towards}. For instance, in \cite{song2021towards} the authors propose encoding structural information in summarization models and in \cite{fan2018robust,chen2022towards}, the authors propose new architectures/models to improve factual robustness of generated summaries. Note that the research question we address in our work is distinct from this line of work as our goal is to identify if minor perturbations of the input space can lead the LLMs to generate different summaries (both the original and paraphrased summaries could still be faithful/factual to the input article). Other past work \cite{chen2020cdevalsumm,lux2020factual,krishna2022improving} has investigated robustness of non-LLM summarizers in distribution shift scenarios by assessing how models trained on a particular dataset perform on other datasets \cite{chen2020cdevalsumm}, analyzing the temporal issue where articles are changing over time \cite{lux2020factual}, and adding synthetic data pipeline / text extraction noise to the input to assess performance \cite{krishna2022improving}. As can be seen, our paper considers a somewhat related but overall separate problem where we only perturb very few sentences to retain semantic similarity (“relevance paraphrasing”), and then observe if that changes the sentence selection process the LLM uses for summarizing and subsequent summary quality. Past work has not utilized relevance paraphrasing in this manner for robustness analysis possibly due to the general brittleness issues of non-LLM summarization models \cite{kryscinski2019neural}.

\section{Human Evaluation}\label{appendix:HumanEval}

The following instructions were provided to the annotators\footnote{The annotators were grad students of UC Davis residing in Davis, California.} (similar protocol as to \cite{zhang2024benchmarking}: \textit{Annotate the summary quality of the following Article-summary pairs using the definitions. Your scores will be presented in a research article. Definitions: }

\begin{itemize}
    \item \textit{\textbf{Faithfulness (Binary Score 0 or 1):} Whether the summary is faithful to the article or not (on topic).}
    \item \textit{\textbf{Coherence (Integer Score between 1 to 5):} Whether the summary is well-structured and organized. Should not just be a heap of related information}
    \item \textit{\textbf{Relevance (Integer Score between 1 to 5): }Whether the summary selects important content from the source. Summary should only include the important info.}
\end{itemize}

The results obtained are as follows. For each dataset-model pair the best summary (in bold) is the one with overall high performance on the 3 sub-metrics (to break ties we use the overall highest scores):

\begin{table}[!ht]
    \centering
    \resizebox{0.45\textwidth}{!}{
    \begin{tabular}{cccc}
    \hline
        \multicolumn{4}{c}{\textbf{Gold}} \\ \hline
        \textbf{Dataset-Model} & \textbf{Faithfulness} & \textbf{Coherence} & \textbf{Relevance} \\ \hline
        Xsum GPT & 0.520(0.242) & 3.390(1.257) & 2.280(0.440) \\
        Xsum Llama & 0.550(0.201) & 2.870(0.640) & 2.380(0.779) \\
        \textbf{Xsum Dolly} & 0.630(0.163) & 3.580(0.984) & 2.820(0.505) \\ 
        \textbf{Xsum Mistral} & 0.630(0.123) & 3.470(0.456) & 3.340(0.560) \\ 
        Reddit GPT & 0.720(0.077) & 2.750(0.622) & 2.820(0.645) \\ 
        Reddit Llama & 0.670(0.133) & 2.950(0.572) & 2.950(0.230) \\ 
        \textbf{Reddit Dolly} & 0.830(0.095) & 3.100(0.272) & 3.620(0.308) \\ 
        \textbf{Reddit Mistral} & 0.790(0.160) & 3.090(1.057) & 3.270(0.500) \\ \hline
    \end{tabular}}
\end{table}

\begin{table}[!ht]
    \centering
    \resizebox{0.45\textwidth}{!}{
    \begin{tabular}{cccc}
    \hline
        \multicolumn{4}{c}{\textbf{Original}} \\ \hline
        \textbf{Dataset-Model} & \textbf{Faithfulness} & \textbf{Coherence} & \textbf{Relevance} \\ \hline
        \textbf{Xsum GPT} & 0.950(0.077) & 4.000(0.514) & 4.380(0.375) \\ 
        \textbf{Xsum Llama} & 0.990(0.030) & 3.520(0.600) & 3.680(0.380) \\ 
        Xsum Dolly & 0.670(0.116) & 3.370(0.552) & 2.670(0.289) \\ 
        Xsum Mistral & 0.770(0.177) & 3.270(0.668) & 2.800(0.608) \\ 
        \textbf{Reddit GPT} & 1.000(0.000) & 4.220(0.816) & 4.400(0.492) \\ 
        Reddit Llama & 0.920(0.101) & 3.850(0.358) & 3.900(0.232) \\ 
        Reddit Dolly & 0.520(0.248) & 2.340(0.534) & 2.070(0.315) \\ 
        Reddit Mistral & 0.400(0.077) & 3.100(0.439) & 2.220(0.477) \\ \hline
    \end{tabular}}
\end{table}

\begin{table}[!ht]
    \centering
    \resizebox{0.45\textwidth}{!}{
    \begin{tabular}{cccc}
    \hline
        \multicolumn{4}{c}{\textbf{Paraphrased}} \\ \hline
        \textbf{Dataset-Model} & \textbf{Faithfulness} & \textbf{Coherence} & \textbf{Relevance} \\ \hline
        Xsum GPT & 0.920(0.073) & 3.760(0.545) & 3.720(0.146) \\ 
        Xsum Llama & 0.990(0.030) & 3.550(0.303) & 3.590(0.305) \\ 
        Xsum Dolly & 0.820(0.186) & 3.370(0.622) & 2.720(0.305) \\ 
        Xsum Mistral & 0.820(0.085) & 3.250(0.648) & 2.870(0.590) \\ 
        Reddit GPT & 0.890(0.030) & 4.130(0.497) & 3.920(0.581) \\ 
        \textbf{Reddit Llama} & 0.950(0.071) & 3.880(0.363) & 4.120(0.093) \\ 
        Reddit Dolly & 0.500(0.179) & 2.230(0.737) & 2.040(0.506) \\ 
        Reddit Mistral & 0.370(0.173) & 2.930(0.509) & 1.950(0.469) \\ \hline
    \end{tabular}}
\end{table}

\section{NLI based Evaluation}\label{appendix:NLI}

 We use MENLI \cite{chen2023menli} to perform NLI evaluation of our summaries as it is a recent work introducing metrics for robust NLI. We sub-sample the CNN, XSum and Reddit datasets to 2k samples each and evaluate on the llama2 outputs. For MENLI, our evaluation parameter settings were: Direction=rh, formula=‘e-c’, nli\_weight=0.3,  combine\_with="BERTScore-F", model=”D” in line with the best performing configuration for summarization in their paper.

\begin{table}[!ht]
\centering
\resizebox{0.225\textwidth}{!}{
\begin{tabular}{cc}
\hline
                   & \multicolumn{1}{c}{\textbf{Llama-2\textsubscript{13B}}}  \\
                   \hline
\textbf{Dataset}   & \textbf{\% change}  \\
\hline
CNN  & {\color[HTML]{9A0000}(-)0.566}               \\

XSum  & {\color[HTML]{9A0000}(-)1.193}               \\

News & {\color[HTML]{9A0000}(-)3.226}                 \\

Reddit & {\color[HTML]{9A0000}(-)23.53}           \\
\hline

\end{tabular}}
\end{table}

We can observe a consistent degradation in the quality of outputs that are in-line with the findings of our paper. The performance change is negative throughout and the drop in performance is highest on Reddit (-23.53\%) and News (-3.266\%). This further emphasizes the effect of our relevance paraphrasing to analyze the robustness of LLM summarization.

\section{Results for Traditional Summarization Models BART and Pegasus}\label{appendix:BART}

We provide robustness results after \textit{relevance paraphrasing} for the \textit{fine-tuned} versions of BART and Pegasus.

\begin{table}[!ht]
    \centering
    \small
    \begin{tabular}{llcc}
        \hline
        \textbf{Dataset} & \textbf{Metric} & \textbf{BART} & \textbf{Pegasus} \\
        \hline
        \multirow{4}{*}{\textbf{CNN}} & Rouge1 & {\color[HTML]{9A0000}(-)13.04} & {\color[HTML]{9A0000}(-)14.1} \\
        & Rouge2 & {\color[HTML]{9A0000}(-)33.71} & {\color[HTML]{9A0000}(-)34.61} \\
        & RougeL & {\color[HTML]{9A0000}(-)18.38} & {\color[HTML]{9A0000}(-)19.73} \\
        & BertScore & {\color[HTML]{9A0000}(-)0.676} & {\color[HTML]{9A0000}(-)0.828} \\
        \hline
        \multirow{4}{*}{\textbf{Xsum}} & Rouge1 & {\color[HTML]{9A0000}(-)3.14} & {\color[HTML]{9A0000}(-)2.978} \\
        & Rouge2 & {\color[HTML]{9A0000}(-)5.601} & {\color[HTML]{9A0000}(-)4.836} \\
        & RougeL & {\color[HTML]{9A0000}(-)3.53} & {\color[HTML]{9A0000}(-)3.139} \\
        & BertScore & {\color[HTML]{9A0000}(-)0.22} & {\color[HTML]{9A0000}(-)0.218} \\
        \hline
        \multirow{4}{*}{\textbf{News}} & Rouge1 & {\color[HTML]{9A0000}(-)10.71} & {\color[HTML]{9A0000}(-)12.3} \\
        & Rouge2 & {\color[HTML]{9A0000}(-)21.9} & {\color[HTML]{9A0000}(-)22.71} \\
        & RougeL & {\color[HTML]{9A0000}(-)11.6} & {\color[HTML]{9A0000}(-)12.64} \\
        & BertScore & {\color[HTML]{9A0000}(-)0.587} & {\color[HTML]{9A0000}(-)0.763} \\
        \hline
        \multirow{4}{*}{\textbf{Reddit}} & Rouge1 & {\color[HTML]{9A0000}(-)5.4} & {\color[HTML]{9A0000}(-)8.81} \\
        & Rouge2 & {\color[HTML]{9A0000}(-)15.41} & {\color[HTML]{9A0000}(-)22.43} \\
        & RougeL & {\color[HTML]{9A0000}(-)6.263} & {\color[HTML]{9A0000}(-)10.15} \\
        & BertScore & {\color[HTML]{9A0000}(-)0.137} & {\color[HTML]{9A0000}(-)0.38} \\
        \hline
    \end{tabular}
    \caption{Robustness degradation after relevance paraphrasing for BART and Pegasus.}
\end{table}

We can clearly see a robustness degradation across the board. For example, BART shows a higher drop in quality than Llama2-13b-chat across all metrics and datasets, except for the ROUGE metrics in the News dataset. However, this is expected, as humans tend to prefer LLM-generated summaries over those produced by supervised models. Hence, poorer robustness is a consequence of poorer overall performance. Therefore, we recommend using LLMs for abstractive summarization in general. However, even LLMs need to be made more robust (i.e., they are not as resilient to relevance paraphrasing as human summarizers) despite their better performance compared to BART and Pegasus.

\section{Successive Prompting of Original Articles}\label{appendix:successive}

We perform successive prompting and compare the results with the results in Table \ref{tab:res}. We can clearly see that there is significantly less change in the outputs and that it is also in not as uniform a degradation. A similar comparison can be made with Table \ref{tab:temp0}.

\begin{table}[!htb]
\centering
\caption{Performance change (\%) observed after prompting the LLM with the same article summary pair twice}
\label{tab:succ}
\resizebox{0.49\textwidth}{!}{%
\begin{tabular}{llllllcl}
\toprule
\multirow{1}{*}{Datasets} & \multirow{1}{*}{Metrics} & \multicolumn{1}{c}{\textbf{Llama-2\textsubscript{13B}}} & \multicolumn{1}{c}{\textbf{GPT-3.5\textsubscript{Turbo}}}  & \multicolumn{1}{c}{\textbf{Dolly-v2\textsubscript{7B}}} & \multicolumn{1}{c}{\textbf{Mistral\textsubscript{7B}}} \\ \cmidrule{3-6}

 &  & \multicolumn{4}{c}{Performance Change (\%)}   \\ \midrule
 & ROUGE-1 & \multicolumn{1}{c}{\color[HTML]{9A0000} (-)0.039} & \multicolumn{1}{c}{\color[HTML]{036400} (+)0.328} & \multicolumn{1}{c}{\color[HTML]{9A0000} (-)0.119} & \multicolumn{1}{c}{\color[HTML]{036400} (+)0.014}  \\
 & ROUGE-2 & \multicolumn{1}{c}{\color[HTML]{9A0000} (-)0.088} & \multicolumn{1}{c}{\color[HTML]{036400} (+)0.909} & \multicolumn{1}{c}{\color[HTML]{036400} (+)0.016} & \multicolumn{1}{c}{\color[HTML]{036400} (+)0.134}  \\
 & ROUGE-L & \multicolumn{1}{c}{\color[HTML]{9A0000} (-)0.013} & \multicolumn{1}{c}{\color[HTML]{036400} (+)0.294} & \multicolumn{1}{c}{\color[HTML]{9A0000} (-)0.040} & \multicolumn{1}{c}{\color[HTML]{9A0000} (-)0.029}  \\
\multirow{-4}{*}{\textbf{CNN}} & BertScore & \multicolumn{1}{c}{\color[HTML]{036400} (+)4.66E-08} & \multicolumn{1}{c}{\color[HTML]{036400} (+)0.009} & \multicolumn{1}{c}{\color[HTML]{036400} (+)2.60E-08} & \multicolumn{1}{c}{\color[HTML]{036400} (+)7.75E-08}
  \\ \midrule

 & ROUGE-1 & \multicolumn{1}{c}{\color[HTML]{9A0000} (-)0.014} & \multicolumn{1}{c}{\color[HTML]{036400} (+)0.347} & \multicolumn{1}{c}{\color[HTML]{036400} (+)0.069} & \multicolumn{1}{c}{\color[HTML]{9A0000} (-)0.033}  \\
 & ROUGE-2 & \multicolumn{1}{c}{\color[HTML]{9A0000} (-)0.033} & \multicolumn{1}{c}{\color[HTML]{9A0000} (-)1.622} & \multicolumn{1}{c}{\color[HTML]{036400} (+)0.049} & \multicolumn{1}{c}{\color[HTML]{9A0000} (-)0.329}  \\
 & ROUGE-L & \multicolumn{1}{c}{\color[HTML]{036400} (+)0.096} & \multicolumn{1}{c}{\color[HTML]{036400} (+)0.128} & \multicolumn{1}{c}{\color[HTML]{9A0000} (-)0.052} & \multicolumn{1}{c}{\color[HTML]{9A0000} (-)0.046}  \\
\multirow{-4}{*}{\textbf{XSum}} & BertScore & \multicolumn{1}{c}{\color[HTML]{036400} (+)9.91E-08} & \multicolumn{1}{c}{\color[HTML]{036400} (+)0.005} & \multicolumn{1}{c}{\color[HTML]{036400} (+)1.33E-07} & \multicolumn{1}{c}{\color[HTML]{036400} (+)7.43E-08}
  \\ \midrule

 & ROUGE-1 & \multicolumn{1}{c}{\color[HTML]{036400} (+)0.146} & \multicolumn{1}{c}{\color[HTML]{036400} (+)1.202} & \multicolumn{1}{c}{\color[HTML]{036400} (+)0.552} & \multicolumn{1}{c}{\color[HTML]{9A0000} (-)0.141}  \\
 & ROUGE-2 & \multicolumn{1}{c}{\color[HTML]{036400} (+)0.530} & \multicolumn{1}{c}{\color[HTML]{9A0000} (-)1.919} & \multicolumn{1}{c}{\color[HTML]{9A0000} (-)0.137} & \multicolumn{1}{c}{\color[HTML]{9A0000} (-)0.343}  \\
 & ROUGE-L & \multicolumn{1}{c}{\color[HTML]{036400} (+)0.218} & \multicolumn{1}{c}{\color[HTML]{9A0000} (-)0.268} & \multicolumn{1}{c}{\color[HTML]{036400} (+)0.431} & \multicolumn{1}{c}{\color[HTML]{036400} (+)0.185}  \\
\multirow{-4}{*}{\textbf{News}} & BertScore & \multicolumn{1}{c}{\color[HTML]{036400} (+)0.000} & \multicolumn{1}{c}{\color[HTML]{036400} (+)0.059} & \multicolumn{1}{c}{\color[HTML]{036400} (+)5.86E-08} & \multicolumn{1}{c}{\color[HTML]{036400} (+)1.17E-07}  \\ \midrule

 & ROUGE-1 & \multicolumn{1}{c}{\color[HTML]{9A0000} (-)0.047} & \multicolumn{1}{c}{\color[HTML]{9A0000} (-)2.392} & \multicolumn{1}{c}{\color[HTML]{9A0000} (-)0.160} & \multicolumn{1}{c}{\color[HTML]{036400} (+)0.032}  \\
 & ROUGE-2 & \multicolumn{1}{c}{\color[HTML]{9A0000} (-)0.375} & \multicolumn{1}{c}{\color[HTML]{9A0000} (-)3.429} & \multicolumn{1}{c}{\color[HTML]{9A0000} (-)0.654} & \multicolumn{1}{c}{\color[HTML]{9A0000} (-)0.757}  \\
 & ROUGE-L & \multicolumn{1}{c}{\color[HTML]{036400} (+)0.035} & \multicolumn{1}{c}{\color[HTML]{9A0000} (-)2.060} & \multicolumn{1}{c}{\color[HTML]{9A0000} (-)0.393} & \multicolumn{1}{c}{\color[HTML]{9A0000} (-)0.004}  \\
\multirow{-4}{*}{\textbf{Reddit}} & BertScore & \multicolumn{1}{c}{\color[HTML]{9A0000} (-)2.60E-07} & \multicolumn{1}{c}{\color[HTML]{036400} (+)0.043} & \multicolumn{1}{c}{\color[HTML]{036400} (+)2.60E-07} & \multicolumn{1}{c}{\color[HTML]{9A0000}{(-)9.00E-08}}  \\ \bottomrule

\end{tabular}%
}\vspace{-2mm}
\end{table}

\section{Code and Reproducibility}\label{code}

We open-source our code and provide it as a Github repository: \url{https://anonymous.4open.science/r/Relevance-Paraphrasing-D1EB/}.

% \url{https://github.com/HadiAskari/Relevance-Paraphrasing}

The repository contains instructions for how to reproduce our results and analyze the findings for each model. All the original summaries and articles, as well as the paraphrased articles and summaries for each model and dataset are also provided in this repository for qualitative analysis. We used Python 3.8.10 for all experiments. The experiments were conducted on Ubuntu 20.04 using NVIDIA GeForce RTX A6000 GPUs running with CUDA version 12.0.

\end{document}